# JAM: The JavaScript Agent Machine for Distributed Computing and Simulation with reactive and mobile Multi-agent Systems - A Technical Report

### Stefan Bosse

### Univ. of Bremen, Dept. of Mathematics & Computer Science, and Institute for Digitization, Bremen, Germany

**Abstract**. Agent-based modelling (ABM), simulation (ABS), and distributed computation (ABC) are established methods. The Internet and Web-based technologies are suitable carriers. This paper is a technical report with some tutorial aspects of the JavaScript Agent Machine (JAM) platform and the programming of agents with AgentJS, a sub-set of the widely used JavaScript programming language for the programming of mobile state-based reactive agents. In addition to explaining the motivation for particular design choices and introducing core concepts of the architecture and the programming of agents in JavaScript, short examples illustrate the power of the JAM platform and its components for the deployment of large-scale multi-agent system in strong heterogeneous environments like the Internet. JAM is suitable for the deployment in strong heterogeneous and mobile environments. Finally, JAM can be used for ABC as well as for ABS in an unified methodology, finally enabling mobile crowd sensing coupled with simulation (ABS).

# Contents

## 1. Introduction

Distributed agent-based computing (ABC) provides some advantages over traditional data processing using central services. Decentralised solving of problems, e.g., energy management [1], can be solved more efficiently than centralised solving and can reduce communication and optionally computational complexity. Most important feature of agent-based systems are their improved resilience and robustness against partial failures. The agent taxonomy addresses three areas: Languages, tools and algorithms, and platforms. A thorough discussion can be found in [2].

To simplify the development and deployment of multi-agent systems in pervasive and ubiquitous applications connected by the Internet and deployed in strong heterogeneous environments including the IoT and mobile networks an unified, portable, and embeddable agent processing platform is required. To achieve scalability and high performance, the agent platform architecture must be closely designed with the agent programming model hand-in-hand. This technical report bases on several scientific publication from the past 10 years [3-19]. All publications addresses the deployment of Multi-agent systems and uses a programmable unified agent processing platform. Initially, the programmable agents were implemented in hardware using FPGA logic [3,4], then a software and hardware implementation of a FORTH stack processor with an extended agent-specific instruction set was introduced (AgentFORTH) [6,7]. Finally, the Web and Cloud deployment gains focus and the JavaScript Agent Machine (JAM) was invented. All different platforms base on the same meta level language AAPL (Activity-based Agent Programming Language) [12]. All agent programming languages used in our past work were dialects or sub-sets of AAPL. The aim of this technical report is the detailed and rigorous description of JAM and the AgentJS programming language processed by JAM with a lot of background information and design patterns. JAM is used in a wide range of applications including distributed machine learning [9], distributed meta material (robotic) control [15] combined with simulation, mobile crowd sensing and social data mining and chat bots [14,16,18], distributed energy management [13], and distributed traffic control [17]. JAM is used in physical systems as well as in simulations (JAM is the simulator).

There are several design choices regarding the agent modelling, the programming of agents, the agent processing architecture, and agent interaction. The design choices depend on the deployment of agents, the fields of application, and most important the physical and digital environment the agents are situated in. The agent behaviour models can range from lower to higher abstraction levels with different expressiveness. The lower levels are more close to traditional data processing architectures and procedural or object-orientated programming, whereas higher abstraction levels focus on knowledge representation, cognitive representation, ontologies, typically using declarative programming models and languages. The Belief-Desires-Intentions (BDI) architecture is a well-known and widely used agent behaviour model. But still traditional imperative and procedural/object orientated languages like Java are used to program agents.

Agents commonly require a virtual machine (VM) acting as an agent processing platform (APP). On one side, the APP has to isolate agents from each other, on the other side, the APP has to provide sufficient agent communication and interaction. The APP must fulfil a core set of features:

1. Portability and independence from the host platform;

2. Isolated and optionally multi-threaded processing of code (containers);

3. Weak coupling of code processing to the underlying API (relaxing hard API matching issues typical in modular software);

4. (a) High performance and (b) low memory requirements;

5. Web and Web browser deployment;

6. Support for IoT and embedded systems.

Java is a widely used programming language and VM. Popular agent frameworks, e.g., Jade [25], rely on Java. But Java only satisfy requirements 1, 2, and partially 6. Web deployment can be basically only achieved with the "native" Web programming language JavaScript. JavaScript (originally named ECMAscript) satisfies requirements 1, 3, 4a, 5 ,6, and partially 2.

Today, a main part of human-machine interaction takes place via the Web Browser. Web browsers are increasingly used for computation and communication, e.g., text processing, Machine Learning, and video conferencing. From a practical point of view, agents should be directly implemented in JavaScript (JS), which is a well-regarded and public widespread used programming language. JS execution platforms are available for a very broad range of devices and operating systems, e.g., Intel x86/x64, Arm32, Linux, Windows, Solaris, MacOS, FreeBSD, Android, IOS, and many more. Furthermore, the implementations of mobile agents directly in JS would benefit from actually existing high-performance JS VMs, e.g., Googles Chrome V8 or Mozillas SpiderMonkey engines with Just-in-Time native code compilation (JIT). At a glance, JS is a very simple but highly dynamic language covering procedural, object-orientated, and functional concepts.

Even if a JIT-based VM is used, full code-to-text and text-to-code transformation is preserved at any execution time, including functions and data. This enables the capability of code morphing at runtime, a prerequisite for AAPL-based agents, used to store the current state of an agent process (e.g., prior to migration) and to modify the behaviour of an agent by applying a re-composition to the ATG by the agent itself. In contrast to JAVA and common JAVA-based agent frameworks (e.g., JADE), JS has a loose coupling to and low dependencies of the underlying execution platform. This is a significant advantage over JAVA or C programming languages, which must be always compiled before the code can be executed, and being very sensitive for API and library mismatches. JS considers functions as first-order values, enabling code reconfiguration on-the-fly like any other data modification using the built-in code evaluation function.

An agent can be considered as a computational unit situated in an environment and world, which performs computation, basically hidden for the environment, and interacts with the environment to exchange basically data. A common computer is specialized to the task of calculation, and interaction with other machines is encapsulated by calculation and performed traditionally by using messages. An agent behaviour can be reactive or proactive, and it has a social ability to communicate, cooperate, and negotiate with other agents. Pro-activeness is closely related to goal-directed behaviour including estimation and intentional capabilities.

Agents record information about an environment state $e \in E$ and history $h: e_0 \rightarrow e_1 \rightarrow e_2...$ Let $I = S \times D$ be the set of all internal states of the agent consisting of the set of control states $S$ related to activities and internal data $D$. An agent's decision-making process is based on this information. There is a perception function *see* mapping environment states on perceptions, a *next* function mapping internal states and perceptions $p \in Per$ on internal states (state transition), and the action-selection function *action*, which maps internal states on actions $act \in Act$.

Actions of agents modify the environment, which is seen by the agent, thus the agent is part of the environment. Learning agents can improve their performance to solve a given task if they analyse the effect of their action on the environment. After an action was performed the agent gets a feedback in form of a reward $r(t) = r(e_t, a_t)$. There are strategies $\pi: E \rightarrow A$ that map environment states on actions. The goal of learning is to find optimal strategies $\pi^*$ that is a subset of $p$.

The JavaScript Agent Machine (JAM) is suitable for the deployment in strong heterogeneous and mobile environments. JAM can eb used for distributed computing as well as for simulation in an unified methodology and with a unified agent programming model, finally enabling mobile crowd sensing coupled to simulation.

In the following section the fundamentals of the agent model, agent adaptivity, agent interaction, and the programming language AgentJS are introduced. After the introduction of the background, JAM itself and its architecture is introduced. An extended section is dedicated to the description of the AgentJS programming language. JAM bases on a modular library concept. Some application programs using the JAM core library are introduced and described. A section dedicated to performance metrics and evaluation is shown and discussed. The simulation environment SEJAM based on JAM is introduced with an extended example project. Finally, the use-case of mobile crowd sensing coupled to simulation is presented. Methodological as well as technical design choices and issues are discussed.

The novelty of JAM is its simple and unified design providing a high degree of portability and a broad range of fields of application. The usage of the widely established JavaScript programming language enables programming of multi-agent systems )MAS) by beginners and non-experts. In addition, MAS can be deployed in the Internet and Web domains, the most prominent human-machine interface. All programming examples can be tested using JAM or the Web JAM laboratory, available on github [26] (distribution and source code). Only a Web browser and the node.js core binary [27] are required.

## 2. The Agent Model

An agent behaviour model can be partitioned into the following tasks, which must be reflected by an agent programming language model by providing suitable statements, types, and structures:

**Computation**

One of the main tasks and the basic action is computation of output data from input data and stored data (history). Principally functional and procedural (or object-orientated) programming models are suitable, but history incorporating computed data and storage is handled only by the procedural programming model consequently.

**Communication**

Communication as the main action serves two canonical goal tuples: (data exchange and synchronization), (interaction with the environment and with other agents). The latter goal tuple can be reduced to agent interaction only if the environment is handled by an agent, too. Communication between agents can link single agents to a Multi-agent system, by using peer-to-peer or group communication paradigms.

**Mobility**

Mobility of agents increases the perception and interaction environment significantly. Mobile agents can migrate from one computing environment to another finally continuing there their processing. The state of an agent, consisting of the control and data state, must be preserved on migration.

**Reconfiguration**

Traditional computing systems get a fixed behaviour and operational set at design time. Adaptation in the sense of behavioural reconfiguration of a system at run-time can significantly increase the reliability and efficiency of the tasks performed.

**Replication**

These are the methods to create new agents, either created from templates or by forking child agents, which inherit the behaviour and state of the parent agent, finally executing in parallel. Replication is one of the major agent behaviours to compose distributed computational and reactive systems.

**Agents and Objects**

Modern data processing is often modelled based on object-orientated programming paradigms. But there is a significant difference between agents and objects. Objects are computational units encapsulating some state and are able to perform actions (by applying methods) and communicate commonly by message passing. Objects are related to object-orientated programming and are not (or less) autonomous in contrast to agents, and they are commonly immobile. The common object model has nothing common with proactive and social behaviour. But agents can be implemented on top of the object model with methods acting on objects. The modification of the behaviour engine is basically not supported by the object programming model. Agents decide for themselves, in contrast objects require external computational units, like operating systems or users, for the decision-making process. Though object-orientated programming can be extended by parallel and concurrent processing (multi-threading), multi-agent systems are inherently multi-threaded.

## 2.1 Activity-Transition Graphs

Agent models can be roughly divided into reactive and deliberative agent behaviour models. The agent behaviour defines

- a set of sensors and beliefs about the world (its data set), basically stored in the agents body variables,

- a set of events that it will respond to (signals, tuples, sensor changes),

- a set of goals that it may desire to achieve, basically implemented by the set of activities, and

- a set of plans that describe how it can handle the goals or events that may arise activating activities.

The behaviour of an activity-based agent is characterized by an agent state, which is changed by activities. Activities perform perception, plan actions, and execute actions modifying the control and data state of the agent. Activities and transitions between activities are represented by an activity-transition graph (ATG). The transitions start activities commonly depending on the evaluation of agent data (body variables), representing the data state of the agent, shown in Fig. 1.

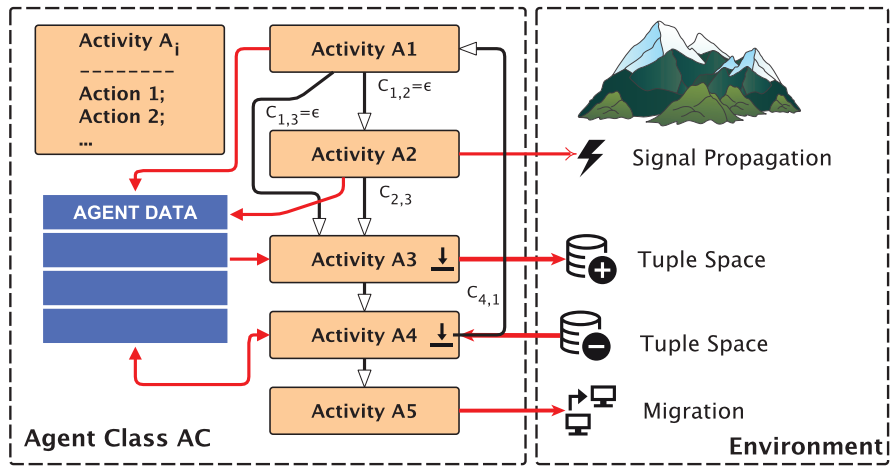

*Fig. 1. Agent behaviour given by an Activity-Transition Graph and the interaction with the environment performed by actions executed within activities*

The agent behaviour is modelled with a cyclic directed Activity-Transition graph (ATG). An activity $a$ from a set of activities $a \in A$ is an active (functional) node of this graph that performs actions, i.e., basically computation, communication, migration, and agent control. Transitions are directed active (functional) edges between nodes. A transition $t$ from a set of transitions $t \in T$ can be conditional and depends on the evaluation of the agent's state. The agent state is mainly contained in the set of private body variables and the control state represented by the *next* activity pointer.

An agent is created from a class constructor $AC_i$ defining the set of activities $A$, transitions $T$, body variables $V$, procedures $P$, and

signal handlers $H$:

$$AC_i = \langle A, T, V, P, H \rangle$$
$$A = \{a_1, a_2, .., a_n\}$$
$$a_i = \{i_1, i_2, ..|i_u \in ST\}$$
$$T = \{t_{ij} = t_{ij}(a_i, a_j, cond)|a_i \xrightarrow{cond} a_j; i, j \in \{1, 2, .., n\}\} \quad (1)$$
$$V = \{v_1, v_2, .., v_m\}$$
$$P = \{p_1, p_2, .., p_i\}$$
$$H = \{h_1, h_2, .., h_j\}$$

An activity $a_i$ consists of sequential statements $\{st_j\}$ from a set of statements $st_j \in ST$. The statement set can be different for different agents and can depend on environmental settings and constraints like the privilege level of the agent.

An agent can be suspended for waiting on events (including time events). Since not all programming languages and processing environments support process control flow blocking, only transition execution and activation be blocked. That means, there may be only one blocking operation inside one activity, e.g., sleeping or waiting on communication data (see tuple space communication, Sec. 2.3.1).

## 2.2 Dynamic Activity Transition Graphs

The ATG introduced in the previous section provided a dynamic state by the agent body variables (data state). Additionally, the agent behaviour can be changed by changing the functional activity set and the transition set (control state). Adaptivity of agents is provided by self-restructuring of the ATG by adding, changing, or removing of activities and transitions, resulting in a dynamic ATG (DATG).

The basic operations applied to a DATG are reduction (R), res-tructuring (S), and extension (X):

$$A : \{a_i\}_{i=1}^n, T : \{t_{i,j}\}_{i,j=1,\ldots,n}$$
$$R : \langle A, T \rangle \rightarrow \langle A', T' \rangle, |A| \geq |A'|, |T| \geq |T'|$$
$$S : \langle A, T \rangle \rightarrow \langle A, T' \rangle, T \neq T', |T| \neq^? |T'|$$
$$X : \langle A, T \rangle \rightarrow \langle A, T' \rangle, T \neq T', A \neq A', |A| \leq |A'|, |T| \neq^? |T'| \tag{2}$$

The basic transformation principles are shown in Fig. 2. The underlying mechanism of code modification (code morphing) depends on the concrete programming language and the VM that processes the agent platform. JavaScript supports directly code-text and text-code transformations and provides therefore first-level code morphing. Beside agent adaptivity at run-time, ATG sub-classing can be used for the creation of child agents with reduced behaviour and finally reduced code size improving the efficiency of agents.

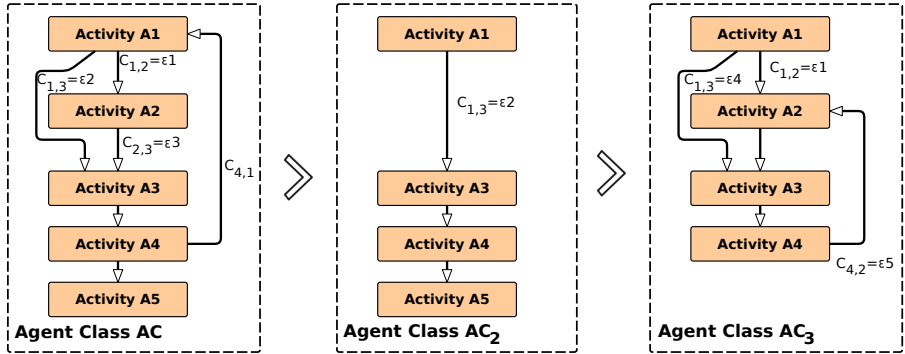

*Fig. 2. Dynamic ATG with modification of activities and/or transitions*

## 2.3  Agent Interaction

Agents are computational processes that are loosely coupled to the underlying processing platform posing low dependence on the programming interface. In consequence, agents are weakly coupled to each other, too. Communication methods that satisfy the weak coupling paradigm are generative tuple spaces [22] and event-based signals, shown in Fig. 3.

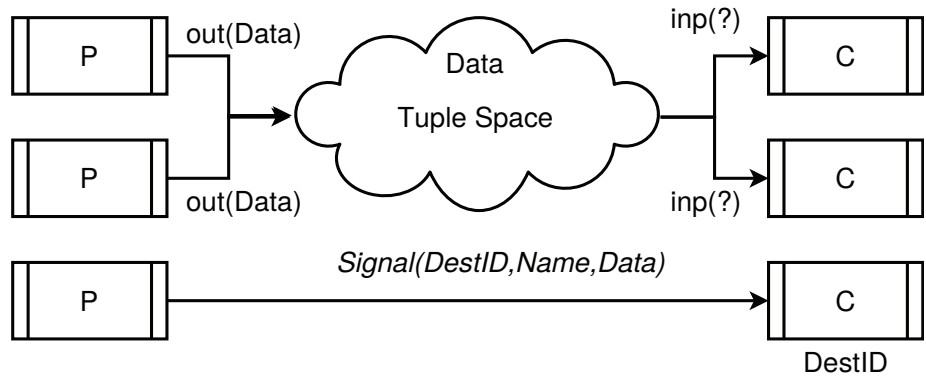

*Fig. 3. Data-driven agent communication using tuple spaces and agent-driven communication using addressed signals*

### 2.3.1  Tuple Spaces

A tuple space is basically a data base that supports producer and consumer operations. A producer process can insert new data by tuples. A tuple is a ad-hoc collection of single data elements, similar to an array. In contrast to arrays, the type signature of a tuple can be heterogeneous, i.e., each element can be any data type, e.g., scalar values like numbers or Boolean values, or aggregated values like arrays or record structures. A tuple $t$ is specified by the number of elements (arity) and the type signature of the tuple:

$$T = \{T_1, T_2, .., T_m\}$$
$$T_i = \{t_{i,1}, t_{i,2}, ..\}$$
$$|t_{i,j}| = i$$
$$t_i = \langle v_1, .., v_i \rangle \tag{3}$$
$$v \in R^n | \text{String} | \text{Array} | \text{Record}$$
$$p_i = \langle x_1, .., x_i \rangle$$
$$x = \perp | v$$

A consumer process can extract tuples (either by removing or by getting a copy) by using patterns ($p$). A pattern is a tuple that consists of actual and formal values. The pattern is applied to all tuples in the space to find matching results. Formal parameters match all data values.

Tuple space communication is anonymous and generative. Producer and consumer processes do not know each other and a tuple can have a longer lifetime than producer process. Tuple space communication poses a lot of advantages, but also disadvantages. The main disadvantage is the initially unlimited lifetime of tuple resulting in an grow and overflow of the tuple space over time. To avoid tuple space flooding, tuples can be assigned a lifetime, too. Either explicitly by the producer process or (and) by the tuple space service. Finding the right lifetime is critical since a too short lifetime misses producer-consumer communication (if the consumer comes too late) or still overflows the tuple space if the lifetime is too large. Therefore, tuple space communication is not guaranteed, which must be considered by the agent behaviour.

Beside the producing operation *out* there are basically two different consuming operations *rd* and *inp*. The input operation removes exactly one matching tuple from the space (atomically), whereas the read operation only returns a copy of the tuple. The consumer operations are synchronised and therefore can block the process control flow of the requesting process (blocking of control flow until a matching tuple was inserted by another process or a timeout occurred).

### 2.3.2  Signals

Signals are simple messages consisting of a signal identifier (number or string name) and can be exchanged by agents. The destination agent must be known (in contrast to tuple space communication). In contrast to uni-cast signals with one specific destination, multi-cast signals can be sent to a group of agents (previously defined), or broadcast signals that can be sent to a all reachable agents. Broadcasting is limited typically to a agents within a given communication range (same node) and limited to a specific agent class.

## 2.4  Software Implementations

The DATG approach can be implemented with a wide variety of programming languages and computer systems, initially implemented with a modified and advanced version of the FORTH programming language and executed on a parallel stack processor system [6]. The initial work had a focus on low-resource embedded systems. To enable the deployment in Web, IoT, and Cloud Internet environments, a successor of this successful low-resource implementation based on the transition from a customised stack-based to a widely used JavaScript programming language with the AgentJS dialect, discussed in the next section.

## 2.5  AgentJS

From the programming language point of view AgentJS is pure JavaScript and conforms fully to the ES2015 standard. But there is a slightly different operational semantic and there are some restrictions. AgentJS describes agent objects and agent constructor functions. In contrast to generic JS classes and objects, there are no free variables and references to higher scopes allowed. A compilation process will eliminate all free and out-of-scope references and binds the `this` variable to the created agent object. The basic template of an agent constructor function conforming to AgentJS is shown in Def. 1.

---

```
1:  function acf(p1?,p2?,..) {
2:    ⎡
3:    ⎢  this.$bodyvar = ε
4:    ⎣
5:    ⎡  this.act = {
6:    ⎢    $act : () => {}
7:    ⎣  }
8:    ⎡  this.trans = {
9:    ⎢    $act : string | function : () → string
10:   ⎣  }
11:   ⎡  this.on? = {
12:   ⎢    $signal : (arg,from) => {}
13:   ⎣  }
14:   ⎡
15:   ⎢  this.next = string
16:   ⎣
17: }
18: function acf(p) : parameter:* → agent object
19: type agent: { $var:*,
20:                act : {function},
21:                trans : {function},
22:                on?: {function},
23:                next:string|object  }
```

*Def. 1. Basic structure of an agent class constructor function*

---

The constructor function defines body variables (must be always initialised, at least with the null object), the activity object containing all activity functions (or lambda expression), the transition object with attribute names equal to the previously defined activities specifying the outgoing activity for a transition, an optional signal handler object, and finally the control state variable *next* that must be point to a valid start activity. A transition section entry is either a static rule providing a fixed next activity (name string or name identifier), or a dynamic transition evaluation function that can return different next activities.

An agent constructor function describes parametrisable agent objects consisting of private body variables, activity functions, transition functions, and optionally event handler functions. The mandatory *next* attribute points to the next activity to be executed and must be initially set to the start activity (*next* can be considered as the agent program counter). The *next* attribute is used to preserve the agent control state on forking or migration, too. A transition attribute from, the *trans* section consists of the start activity name (that must be defined in the *act* section) and a destination activity (also defined in the *act* section). The transition can be unconditional (only one static destination activity is possible) or conditional (more than one destination activity is possible). Conditional transitions require resolution function returning the name of the destination activity. If the agent constructor function is compiled from a string text, activity names can be provided as identifiers, otherwise they must be handled as strings.

Agent activity, transition, and handler functions are always called in the agent object context (the *this* variable always points to the agent object). All functions can use the entire operational AIOS function set with respect to the current privilege level of the agent. Blocking operations may not be chained or nested or called inside loops. Only one blocking operation in an activity function is allowed. Blocking only inhibits activity transition of an agent.

Agent objects are instantiated from the agent constructor function with an individual set of parameters $\{p_i\}$. As there are no free variables, the constructor function parameters must be assigned directly to body variables. An example of an agents constructor function for a simple hello world agent is shown in Ex. 1. All body variables must be initialised by an assignment of a static or dynamic value from the function parameters, and initially undefined variables require the assignment of the null object.

```
1:  function hello(config) {
2:    this.message = config.message;
3:    this.time = 0;
4:    this.delay = config.delay;
5:    this.data = null;
6:    this.act = {
7:      start : () => {
8:        log('Hello world!');
9:        this.time=time();
10:       sleep(this.delay);
11:     },
12:     talk : () => {
13:       log(this.message);
14:       log('I sleep '+(time()-this.time)+' ms');
15:       this.time=time();
16:       sleep(random(10,100));
17:     },
18:     end : () => {
19:       log('I am terminating...')
20:       kill()
21:     }
22:   }
23:   this.trans = {
24:     start : talk,
25:     talk  : () => {
26:       return random(0,100)<50?"end":"talk"
27:     },
28:   }
29:   this.next = "start"
30: }
```

*Ex. 1. Example of an AgentJS constructor function*

## 3. JavaScript Agent Machine

The JavaScript Agent Machine (JAM) is the central part of the multi-agent system framework addressed in this work. In addition to the platform features discussed in the introduction, an agent process platform must satisfy the following constrains from a software design perspective:

1. Portability with low hardware and operating system dependencies;

2. Possibility for deployment on low-resource platforms;

3. Embeddable in any software application including Web pages;

4. Scalability with respect to number of agents and computational agent complexity.

JAM is entirely programmed in JavaScript and satisfies immediately the first and third constrains. The satisfaction of the second and fourth constraints depends on the underlying JavaScript Virtual Machine (JVM). But even the node.js JVM can be deployed on embedded and mobile systems like smartphones (typical requiring at least 1000 MIPS CPU power and 50MB main memory for satisfying responsiveness and scalability).

## 3.1 Agent Worlds: Virtual and Physical Nodes

JAM provides a physical platform for agent processing. A physical JAM node includes the following components and capabilities:

- Agent process scheduler;

- Tuple space data bases;

- Protection and security;

- Agent management and modification;

- Communication including signal and agent mobility.

Beside the physical node, there is a virtualisation layer providing virtual nodes all processed by the physical node. Each JAM instance has at least one virtual node and agents are associated to virtual nodes. The virtual node concept is used for simulation, discussed later. AMP communication ports (and links) are always attached to virtual nodes. This enables virtual networks with virtual circuit links inside a physical node world, shown in Fig. 4. All virtual nodes have individual tuple spaces and process tables, but share the same agent process agent scheduler. Therefore, virtual worlds do not pro-vide speed-up by parallelisation, in contrast, to JAM clusters, which can be easily created by the *jamsh* program. In a JAM cluster, each JAM node is processed by its own VM instance.

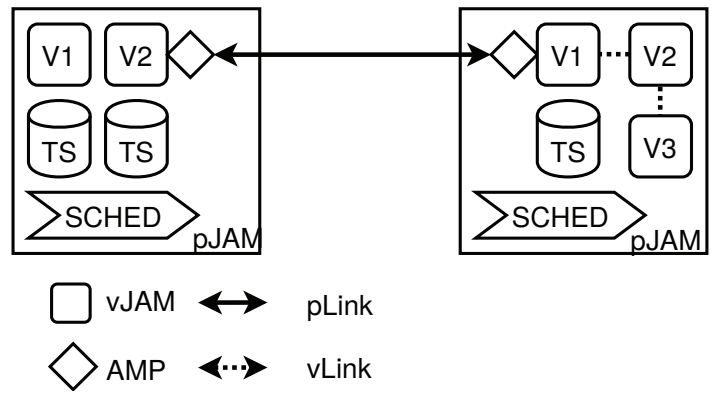

*Fig. 4. Virtual and physical nodes*

## 3.2 Agent Communication

Agents are loosely coupled processes that interact typically by messaging. JAM offers agents tuple spaces and signals for inter-agent communication and in some cases agent-platform communication. Tuple spaces provide connectionless, anonymous, non.addressedn, and generative communication, whereas signals provide message-based addressed communication between agents. Tuple space communication does not require that the agents know from each other, in contrast, to signals. An agent is identified in JAM by a node-unique randomly generated identifier string (typically consisting of eight characters).

Agents can exchange data tuples if they are processed by the same physical/virtual node (local tuple space scope). Each JAM virtual node (each physical node has at least one virtual node) provide a tuple space for tuple of arity 1 up to 10 (the upper number of tuple elements can be extended by the host application). Each single tuple arity is handled independently since tuples can be only ac-cessed by patterns that must have the same number of elements (arity). The node tuple spaces can be accessed by the host application (embedding JAM or the stand-alone JAM shell), too. The host application can provide tuple consumer handlers that are called by the tuple space API if  a new tuple is inserted. Read requests can be passed to producer handler that can create new tuples on request.

Signals are originally used between agents processed on the same (virtual) node. But signals can be propagated to remote nodes, too (or other virtual nodes of a physical node), if, and only if, there is a migration trace between the sender and receiver agents. That means, two agents can exchange signals remotely if they were processed on the same node some time ago. This is always true for parent-child groups. If an agent leaves a node on migration, the current node will remember the agent id and the destination node with the outgoing network link in a temporary cache table. Each time a signal is propagated this path, the cache table is updated. A garbage collector will clean-up traces if they are not used for some time.

Tuple spaces of remote linked nodes can be accessed by using special tuple request signals, used basically for tuple duplication or forwarding.

### 3.3  Agent Input-Output System (AIOS)

The AIOS is the central bridge between the JAM platform or host applications and the agent processes, shown in Fig. 5. The AIOS provides an advanced agent scheduler that is used to schedule all agents of a physical node. It is used for the SEJAM simulator, too. The scheduler performs signal and timer handling, activates agents, call activity and transition functions, and performs error handling, as introduced in Sec. 3.4.

The AIOS provides sets of operations and functions that can be accessed by an agent depending on the privilege level of the agent (AgentJS API). Each privilege level uses its own set of operations. For example, a level zero agent cannot access tuple spaces and cannot migrate. Level three agents are immobile and therefore do not have a `moveto` function in there operational set, too:

$$OP = \{op_0, op_1, op_2, op_3\}$$

$$op = \left\{ \begin{array}{c} \text{math} \\ \text{objects} \\ \text{iterators} \\ \text{agents} \\ \text{control} \\ \text{mobility} \\ \text{negotiation} \\ \text{communication} \\ \text{aux} \end{array} \right\} \qquad (4)$$

$$ag_i^l = ag_i(l, op_l)$$

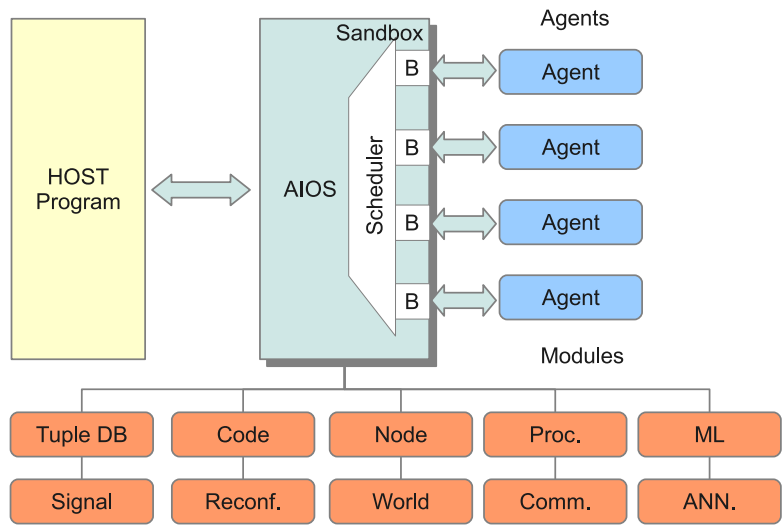

*Fig. 5. Basic AIOS architecture providing a bridge for the agent, host platform, and modules API*

The auxiliary set provides access to helper modules, e.g., Machine Learning or data base access (SQL). The host application can extend each AIOS op-level set by additional functions and objects. For example, the simulator SEJAM will extend the agent op set by simulation specific operations.

### 3.3.1 Code and Data Serialization with JSON+ and JSOB Representation

A textual representation is used as a serialized data and code interchange format, which is a prerequisite for data and code processing in distributed strong heterogeneous platform and network environments, mixing big- and little endian machines, different data word sizes, and data coding. Although, byte-code based interchange formats are widely used, they require a strict compliance of the coding between a sender and a receiver. At any time, a JS object can be converted to text in JSON format at run-time. Originally, JSON was introduced for portable exchange of JS data objects in a textual representation only, being much more compact and easier to interpret than XML. But JSON is limited to data elements, only. References to deeper objects (records or arrays) are always resolved and expanded. Therefore, JSON cannot represent recursive data structures. Although, JS considers functions as values of first order, traditional JSON cannot represent and encode functions. For this reason, the JSON+ format was introduced that supports function encoding as well as special data objects like compact typed arrays and buffers.

A JAM/AgentJS agent is basically a JS object containing data (values, data objects, arrays) and functions, representing the agent activities and transitions of the ATG, requiring an extended JSON text formatter and parser supporting functions, which was introduced in JAM. An entire agent process can be converted at any time to the textual representation (JSON+) preserving its current control and data state, which can be exchanged by different network and agent platform nodes, and that is finally back converted to JS code.

Functions and binary buffers are textual encoded with a base64 code and a tag prefix (e.g., `_PxEnUf_` for function code). The only existing limitation are circular (self) references inside of an object, which still cannot be handled, but not being a real restriction.

Transferring text instead of binary code results in a significantly increased communication cost on agent migration, but the text can be compressed reducing the size significantly (experiments showed that LZ compressing reduces the JSON+ text size and hence the communication costs about 5-6 times). Embedded devices can utilize hardware compressor modules, e.g., using FPGA-based co-processors, maximizing communication efficiency without additional CPU costs.

In addition to the JSON+ format, a customised JavaScript Object (JSOB) format is supported, too. JSOB is a textual representation of a JS object, too, but do not encode functions and is more compact, shown in Ex. 2.

```
ag = {
  x:1, y:2, z:[10,20,30],
  act : {
    main : () => { this.x++ },
    end  : () => { kill() }
  },
  trans : {
    main : end,
  }
  next : main
}
———— JSON+.serialize(ag) ————
{"x":1,"y":2,"z":[10,20,30],"act":
{"main":"_PxEnUf_ZnVuY3Rpb24gKCkgeyB0aGlzLngrKyB9",
"end":"_PxEnUf_ZnVuY3Rpb24gKCkgeyBraWxsKCkgfQ=="},
"trans":{"main":"end"},"next":"main"}
———— JSOB.serialize(ag) ————
{'x':1,'y':2,'z':[10,20,30],'act':
{'main':()=>{ this.x++ },'end':
()=>{ kill() }},'trans':
{'main':"end"},'next':"main"}
```

*Ex. 2. Example of agent object serialisation using JSON+ or JSOB format*

### 3.3.2 The AIOS Sandbox Environment

Stability and robustness of the agent processing platform is one major challenge in the design. Agents can be considered as autonomous or semi-autonomous processes and execution units. But this autonomy requires strictly bound and safe platform environments for the execution of agent processes from arbitrary sources, and the strict isolation of agent processes from each other. At least inference by accident must be excluded. An agent platform must be capable to execute hundreds and thousands of different agent processes. Although, there are extension modules for some JS VMs (e.g., Web worker processes) allowing the execution of a JS program in a separate host process (or thread), these coarse-grained methods are not portable and it creates significant overhead in time and memory space. This is related to start-up time of threads and the fact that each thread creates a new VM instance with limited inter-process communication, required by the AIOS.

Unfortunately, JS has only a very limited and coarse-grained scope context hierarchy, basically limited to function closures and the *this* object (in contrast to Lua, which supports function execution in a freely defined and limited scope), and with one global space shared by all imported modules and evaluated code. This limitation initially prohibits the safe and interference-free execution of multiple agent processes within one JS VM instance. To satisfy the inference constraint, the JS `with (mask) {code}` statement can be used to execute the code with an overlaid name space given by the mask object argument. Typically used to map bound variables on free variables for a limited code scope, here the mask is extended by all global variable names assigned to null value and function name attributes overlaying the global reference scope.

Encapsulating the agent object and its function processing inside the `with` statement can only mask out globally visible objects and functions, but not free variables and functions visible on code evaluation. I.e., each agent instantiation from an agent class constructor function requires additionally a full code-text-code serialisa-tion and deserialisation in a masked context resulting in high computational costs if applied for each agent. A more suitable approach is the sandbox-restriction application to the agent constructor function itself (one time), i.e., the agent is created from a constructor function that was created in a masked environment that masks all undesired global and free variables by null references and by code-text-code serialisation of the constructor function without any side effects. The first implementation choices using the `with` statement are shown in Alg. 1.

```
1: mask      = ∀ $key ∈ global { $key:null }
2: context = ∀ $key ∈ Core { $key:global($key) } ∪
3:              ∀ $key ∈ AIOS(level) { $key:AIOS($key) }
4: maskAndContext = mask ∪ context
5: type sandBox1: object×string → constructor function
6: function sandBox1(maskAndContext, acf) {
7:    var F
8:    eval('with(maskAndContext) { "use strict"; F='
9:               + acf + ' }');
10:   return F
11: }
12: type sandBox2: object×object×string → constructor function
13: function sandBox2 (mask, context, acf) {
14:   var _mask = keys(mask).join(',');
15:   var F =
16:      new Function(_mask,
17:                   '"use strict"; with(this) { f=('+
18:                   acf+').bind(this)} return f')
19:                   .bind(context);
20:   return F();
21: }
22: var acbound1 = sandBox1(context,acf),
23:     acbound2 = sandBox2(mask,context,acf);
24: var ag1 = new acbound1(parameters),
25:     ag2 = new acbound2(parameters);
```

*Alg. 1. First two implementation choices for the sandbox restrictions of the agent constructor function using the* `with` *scope statement compiled by the generic* `eval` *statement or by using the* `Function` *constructor. The agent constructor function acf must be serialised to text before applying the sandbox to remove any free bindings.*

The usage of the `with` statement is deprecated and can introduce a significant performance degradation of the executed agent code (each variable or function access requires the traversal of a long mask table). Another approach transforming the not context-constrained agent class constructor function to the scoped sandbox-restricted constructor function using the `Function` constructor with some expression magic and by splitting the agent name space in a masked global environment (*mask*) and the AIOS object *context* is shown in Alg. 2. This approach avoids the usage of the `with` statement. Instead, the masked variable attributes are passed as function parameters to the scope-restricted agent constructor function evaluation and the desired context variables by a bind process.

```
1: mask    = ∀ $key ∈ global { $key:null }
2: context = ∀ $key ∈ AIOS(level) { $key:AIOS($key) } ∪
3:           ∀ $key ∈ Core { $key:global($key) }
4: type sandBox3: object×object×string → constructor function
5: function sandBox3 (mask, context, acf) {
6:   var pars = keys(context).concat(keys(mask)),
7:       args = pars.map(function (key) { return context[key] });
8:   pars.push('return '+acf);
9:   var foo = new (Function.prototype.bind.apply(Function,pars));
10:   return foo.apply(this,args);
11: }
```

*Alg. 2. Third implementation choice for the agent constructor sandbox application using the* Function *constructor statement only. The mask is passed via the function argument array.*

For each AIOS privilege level and for each agent class *ac* there is one sandbox-restricted constructor function.

The masking approach cannot limit the name space scope (scopes are chained, and higher scopes like the global one are still visible) of the agent process and its functions, but it can be used to override higher scope level and global name qualifiers, and to invalidate references to free variables and functions without compromising other agent processes or the JAM modules. There is still one major vul-nerability introduced by extendible objects, e.g., the *Array* object.

So finally the agent process execution is an execution of a function with a strictly limited operational name space without any bindings to external and free variables and functions. Core objects like *Array* and *String* are replaced by non-expandable proxy versions. Migration of agents is performed by object-to-text serialisation and text-to-object deserialisation. To ensure a sandbox-restricted agent object after migration (no constructor function is involved), the JSON+ parsing and evaluation is always performed inside the `with` statement with a mask environment only providing a selected AIOS set of objects and functions, depending on the agent privilege role, discussed in Section 3.6. A creation of a new agent (from a fork or migration operation) will always first stringify the agent object, and finally coding back a sandbox-restricted agent object free of any free and global object references, which can be executed the AIOS agent scheduler without any interference with the platform and other agents. This approach protects the agent execution and JAM at least against failures by accident using common JS coding styles. The capability of full intrusion protection depends on the JS VM environment itself.

Although, JS supports control flow scheduling by using asynchronous functions (defined with the *async* keyword) and the *await* statement operator, the usage must be prevented since it can create background functions beyond the control of the AIOS (as well as internal background processing using built-in *setTimeout* or *setInterval* functions). AIOS rejects code using asynchronous functions or the await statement operator, and the timer control functions are masked out. The only allowed scheduling point is at the end of an activity function or an element of a scheduling block function.

### 3.3.3 Anonymous Functions and Lambda Expressions

The ATG is composed of anonymous functions. JS supports since the ES6 standard lambda expressions, too, in form of arrow functions. The arrow functions are much more comfortable in program-ming, and especially beginners tend to prefer arrow functions over anonymous functions. But in contrast to other programming languages, anonymous (nameless) functions and arrow functions are not the same in JS. A JS function supports inherently the `this` variable, i.e., each function has its own `this` object variable. Arrow function do not have their own `this` object, instead referencing to the current `this` object of the surrounding context if they are created (compiled). This is a show-stopper on first level for AgentJS since all first- and second-level functions referencing always the agent `this` (self) object. This is achieved by dynamic binding of the agent self-reference on function calls of to the respective agent functions. This is not possible with arrow functions (they ignore any *bind* or *call* mapping of `this`). All arrow functions must be bound to the agent self-object reference on creation. If an agent object is instantiated from a constructor function all arrow functions automatically referencing the right `this` object. But after text serialisation and deserialisation (e.g., on process migration), this is an chicken-and-egg problem. Typically, the `eval` function is used to compile functions from text (or alternatively using the *Function* constructor). But when an arrow function is created, the agent object itself is not available! Instead, an initially empty agent object must be used as a bound `this` reference, finally getting the content from the temporary agent object created by the `eval` function application, shown in Alg. 3.

```
1: function toCode(source,mask) {
2:    // arrow functions do not have a this variable
3:    // they are bound to this in the current context!
4:    // need a wrapper function and a temporary this object
5:    var self={},temp;
6:    function evalCode() {
7:      var code;
8:      try {
9:        // execute script in private context
10:       with (mask) {
11:         eval('"use strict"; code = '+source);
12:       }
13:     } catch (e) { console.log(e) };
```

```
14:    return code;
15:  }
16:  // bind evalCode to empty self == this object
17:  temp = evalCode.call(self);
18:  // arrow functions are now bound to self,
19:  // update self with temp. agent object first-level attributes
20:  self=Object.assign(self,temp);
21:  return self;
22: }
```

*Alg. 3. Arrow function safe agent deserialisation with a masked context from JSOB or expanded JSON+ format*

### 3.4 Agent Scheduling and Check-pointing

JS has a strictly single-threaded execution model with one main thread, and even by using asynchronous callbacks, these callbacks are executed only if the main thread (or loop) terminates, i.e., leaving the control path. This is the second hard limitation for the execution of multiple agent processes within one JS JAM platform and instance. The JAM requires strict control over agents at any time. Agents can migrate between JAM nodes, e.g., connected by the Internet. No agent may compromise a JAM node or affecting the processing of other agents including unlimited run-time. Scheduling in modern JavaScript VMs can be easily achieved by using asynchronous functions with promises and executing them by using `await` statements. But this attractive and simple-to-use control enabling scheduling of agents with a time-multiplexed execution model, this method violates the requirement of full agent process control by the JAM. An example should demonstrate this creating a zombie process. Assume one agent want to sleep (within in activity function) for a specific amount of time with a control path scheduling, shown in Ex. 3. Instead, an AIOS sleep function is used to push the agent processing on a wait queue, finally woken up by the AIOS scheduler by performing the transition to the next activity. But this simulated kind of agent process control limits the number of blocking operations within one activity to one (and blocking occurs only after the activity function returns). Calling of blocking operations in loops require micro scheduling blocks and loop transformations.

```
 1: $act : async () => {
 2:    ...
 3: ×   await asleep(tmo);
 4:    ...
 5: }
 6: async function asleep(tmo) {
 7:    return new Promise((resolve,reject) => {
 8:       setTimeout(resolve,tmo);
 9:    }
10: }
11: $act : () => {
12:    ...
13:    sleep(tmo)
14:    // Scheduler do:
15:    //   now: → agent.blocked →
16:    //           agent.timeout = now+tmo  → ...
17:    //   now+tmo: → !agent.blocked → next(trans)
18: }
```

*Ex. 3. A do-not-do scheduling of an agent using promised asynchronous functions*

If the agent waits for the fulfilment of the promise by the background timer, the AIOS scheduler could execute another agent. But if in any case the sleeping agent was terminated, e.g., explicitly by a kill operation or due to resource violation, the sleeping agent process persists (there is no way to resume sleeping background processes) and a zombie process is created that can continue. For this reason, any promise construction and asynchronous functions must be masked out by the sandbox-restricted context.

Agents processes are scheduled on activity level, and a non-terminating agent process activity would block the entire platform (JS processing strictly single-threaded with one control flow) . Current JS execution platforms including VMs in Web browser programs provide no reliable watchdog mechanism to handle non-terminating JS functions or loops. Although some browsers can detect time-outs, they are only capable to terminate the entire JS program, not a function call. To ensure the execution stability of the JAM and the JAM scheduler, and to enable time-slicing, check-pointing must be injected in the agent code prior to execution if there is no watchdog support built in the VM. This step is performed in the code parsing phase by injecting a call to a checkpoint function CP() at the beginning of a body of each function contained in the agent code, and by injecting the CP call in loop conditional expressions. Though this code injection can reduce the execution performance of the agent code significantly, it is necessary until JS platforms are capable of fine-grained check-pointing and thread scheduling with time slicing. On code-to-text transformation (e.g., prior to a migration request), all CP calls are removed.

Using the node.js VM platform, native code modules executing functions watching the runtime in an Isolate container can be added that enable timeout detection of agent processing (of activity, transition, and signal handler function calls). If the scheduling slice time of an agent is reached, the called functions will throw an exception (SCHEDULE) that is handled by the AIOS agent scheduler. The watchdog modules are detected automatically. If there is no native watchdog support, the software checkpoint injection is used instead.

AIOS provides a main scheduling loop. This loop iterates over all logical nodes of the logical world, and executes one activity of all ready agent processes sequentially. If an activity execution reaches the hard time-slice limit, a SCHEDULE exception is raised, which can be handled by an optional agent exception handler (but without extending the time-slice). This agent exception handling has only an informational purpose for the agent, but offers the agent to modify its behaviour. All consumed activity and transition execution times are accumulated, and if the agent process reaches a soft run-time limit, an EOL exception is raised. This can be handled by an optional agent exception handler, which can try to negotiate a higher CPU limit based on privilege level and available capabilities (only level-2 agents). Any ready scheduling block of an agent and signal handlers are scheduled before activity execution. The scheduling loop is executed iteratively based on event times, i.e., the next scheduler run is performed after a calculated time based on pending events (e.g., sleeping agents) and ready agents.

After an activity was executed, the next activity is computed by the schedule that calls the corresponding transition function in the transition section of the agent.

In contrast to the AAPL model that supports multiple blocking statements (e.g., IO/tuple-space access) inside activities, JS is incapable of handling any kind of process blocking (there is no process and blocking concept). For this reason, scheduling blocks can be used in AgentJS activity functions handled by the AIOS scheduler. Blocking AgentJS functions returning a value use common callback functions to handle function results, e.g., `inp(pat, function(tup) {..})`.

A scheduling block consists of an array of functions (micro activities), i.e., `B(block)` = `B([function(){..}, function(){..},...])`., executed one-by-one by the AIOS scheduler. Each function may contain a blocking statement at the end of the body. The this object inside each function always reference the agent object. To simplify iteration, there is a scheduling loop constructor `L(init, cond, next, block, finalize)` and an object iterator constructor `I(obj, next, block, finalize)`, used, e.g., for array iteration.

Agent execution is encapsulated in a process container handled by the AIOS. An agent process container can be blocked waiting for an internal system-related IO event or suspended waiting for an agent-related AIOS event (caused by the agent, e.g., the availability of a tuple). Both cases stop the agent process execution until an event occurred.

The basic agent scheduling algorithm is shown in Algorithm 8.1 and consists of an ordered scheduling processing type selection, i.e., partitioning agent processing in agent activities, transitions, signals, and scheduling blocks. In one scheduler pass, only one kind of processing is selected to guarantee scheduling fairness between different agents. There is only one scheduler used for all virtual (logical) nodes of a world (a JAM instance). A process priority is used to alternate activity and signal handling of one agent, preventing long activity and transition processing delays due to chained signal process-ing if there are a large number of signals pending.

```
∀ node ∈ world.nodes do
 ∀ process ∈ node.processes do
  Determine what to do with prioritized conditions:
    Order of operation selection:
    0. Process (internal) block scheduling [block]
    1. Resource exception handling
    2. Signal handling [signals]
       - Signals handled if process priority<HIGH
       - Signal handling increase process priority
         temporarily to allow low-latency activity
         and transition function scheduling!
    3. Transition execution
    4. Agent schedule block execution [schedule]
    5. Next activity execution
       - Lowers process priority
  if process.blocked or process.dead or
   process.suspended and process.block=[] and
   process.signals=[] or
   process.agent.next=none and process.signals=[] and
   process.schedule=[]
     then do nothing
  elseif not process.blocked and process.block≠[]
     then execute next block function
  elseif agent resources check failed
     then raise EOL exception
  elseif process.priority < HIGH and process.signals≠[]
     then handle next signal, increase process.priority
  elseif not process.suspended and process.transition
     then get next transition or
          execute next transition handler function
  elseif not process.suspended and process.schedule≠[]
     then execute next agent schedule block function
  elseif not process.suspended
     then execute next agent activity and
       compute next transition,
       decrease process.priority
```

*Alg. 4. JAM agent scheduler algorithm*

### 3.5  Scheduling Blocks

If there is more than one blocking operation in an activity or within a loop statement, or there are computational long running code parts, scheduling blocks must be used. A scheduling block partitions the computational work in sequences or loops of micro functions. Each micro function can call one blocking operation. Although, scheduling blocks can be composed of nested blocks, all blocks are flattened to one linear list.

```
 1:  B([
 2:    function,
 3:    function,
 4:    ..
 5:  ])
 6:  L(
 7:    init:function,
 8:    cond:function()->boolean,
 9:    next:function(),
10:    block:function [],
11:    finalize?:function
12:  )
13:  I(
14:    data:{}|[],
15:    next:function (element),
16:    block: function [],
17:    finalize?:function
18:  )
```

*Def. 6. Templates of AgentJS scheduling block constructor functions that can be used inside agent activity functions. B:Linear block, L:Loop, I:Iterator.*

### 3.6  Agent Roles

Security is another major challenge in distributed systems, especially concerning mobile agent processes. Each agent platform node (i.e,

one physical VM, with multiple JAM VMs operating on the same network node) can receive agents originating either from inside trusting node networks or coming from not trustful networks unknown by the VM. Generally, the VMs have no information about other network nodes except a sub-set of network connectivity used to receive and propagate agent code. To distinguish at least trustful and not trustful agents, different agent privilege levels were introduces, providing different AIOS API sets.

### 3.6.1 Privilege Levels

For security reasons and to limit Denial-of-Service attacks, agent masquerading, spying, or other misuse, agents have different access levels (roles). There are four levels:

---

| | |
|---|---|
| **0** | Guest (not trusting, not mobile) |
| **1** | Normal (maybe trusting, mobile) |
| **2** | Privileged (trusting, mobile) |
| **3** | System (trusting, locally, not mobile) |

---

Privilege level 0 is the lowest level and grants agents only computational and some tuple-space IO statements. The lowest level does not allow agent replication, migration, or the creation of new agents. Level 0 agents have to negotiate their privilege level or rights by using a security capability that grants certain rights or level upgrades.

Level 1 agents can access the common AIOS API operations and capabilities, including agent replication, creation, killing, sending of signals, and code morphing. Level 2 agents are additionally capable to negotiate (set) their desired resources on the current platform, i.e., CPU time and memory limits.

An agent of level n may only create agents up to level n. The highest level (3) has an extended AIOS operation set with host plat-form device access capabilities. Level-3 agents cannot migrate to another node, they are fully stationary. Agents can negotiate resources (e.g., CPU time) and a level raise secured with a capability-key that defines the allowed upgrades. The system level can not be negotiated.

Level-2 agents can initially only be created inside the JAM. They can fork level-2 agents, but after a migration the destination node decides about the privilege level and can lower it, e.g., considering the agent source being not trustful. A migrated agent can get a higher privilege level by negotiation, requiring a valid platform capability with the appropriate rights. After migration, the privilege is lost and must be re-negotiated on a new platform using capabilities. The JAM platform decides the security level. The capability is node specific. A group of nodes can share a common key (identified by a server port). A capability consists of a server port, a rights field, and an encrypted protection field generated with a random port known by the server (node) only and the rights field.

Among the AIOS level, other constrain parameters can be negotiated:

- Scheduling time (maximal time slice for one activity execution, default is 200ms)

- Run time (accumulated agent execution time, default is 5s)

- Living time (overall time an agent can exist on a node before it is killed, default is unlimited)

- Tuple space access limits (number of access operations)

- Memory limits (practically fuzzy, cannot be measured exactly and monitored in JS, usually the entire size of the agent code including private data, typically limited to 50kB serialised code)

Resource constraints are monitored by JAM. Since JAM is executed by generic JS VMs control of memory consumption of agents during the execution of an activity is not possible (inherently not possible by the JS programming paradigm and the automatic memory management fully hidden from the programming level). Although, bound (body) variables could be monitored by proxies, free variables are out of reach by any resource control. Only allocation accessed via the AIOS can be controlled by JAM, e.g., tuple space access.

### 3.6.2 Security with Capabilities

Capabilities are keys to enable specific agent API functions like migration. A capability consists of a service port, a rights bit field, and an encrypted protection field generated with a random port known by the server (node) only and the rights field. The rights field enables specific rights. e.g., the right to negotiate a higher privilege level. A capability is created by the respective service (e.g., the platform itself) by encoding the rights field in a security port using a private port and a one-way function, shown in Fig. 6. Major issues are the secret handling of capabilities on agent migration (requiring secured channels) and passing the right capability to an agent. A broker server is commonly required to distribute and pass capabilities to agents based on classical authorisation and authentication.

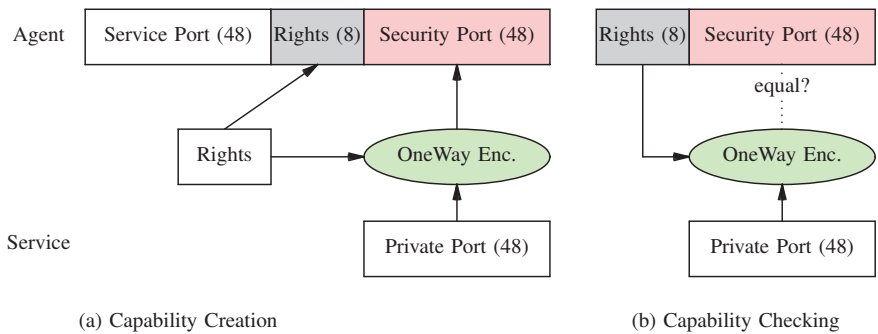

(a) Capability Creation                    (b) Capability Checking

*Fig. 6. (Top) Capability format (a) Capability creation by encoding private security port and new rights field (b) Capability checking by encoding requested rights field and private security port and comparison with provided security port (based on [21])*

Capabilities enable fine-grained control of operations that can be performed by agents and allow loosely coupled self-organising systems, e.g., participation in chats. In self-organising systems the identity of agents is not of primary interest. Instead the actions they can perform is of primary interest.  # Capabilities can be represented in a human readable text format similar to IPv6 addresses and based on Amoeba capability format and concepts [21]:

```
[SS:SS:SS:SS:SS:SS](OBJ(RIGHTS)[PP:PP:PP:PP:PP:PP]
```

The `SS` and `PP` digits are hexadecimal numbers, `OBJ` and `RIGHTS` are decimal numbers. The service port `SS:` defines the target service of the capability, e.g., the resource negotiation service or a communication channel service (AMP). The protection port `PP:` is used to validate the most sensitive rights field of the capability. Commonly, rights are encoded in weighted binary code enabling combinations of rights, e.g., negotiation of CPU resources and AIOS level upgrade. Typical rights bits used by the AIOS are shown in Def. 3.

```
enum Rights = {
  HOST_INFO       : 0x01, HOST_READ      : 0x02,
  HOST_WRITE      : 0x04, HOST_EXEC      : 0x08,
  PSR_READ        : 0x01, PSR_WRITE      : 0x02,
  PSR_CREATE      : 0x04, PSR_DELETE     : 0x08,
  PSR_EXEC        : 0x10, PSR_KILL       : 0x20,
  PSR_ALL         : 0xff,
  NEG_SCHED       : 0x08, NEG_CPU        : 0x10,
  NEG_RES         : 0x20, NEG_LIFE       : 0x40,
  NEG_LEVEL       : 0x80,
  PRV_ALL_RIGHTS  : 0xff
}
```

*Def. 3. AIOS rights bits protecting sensitive agent operations (like negotiation with the agent platform); HOST: platform, PSR: Process control, NEG: Negotiation*

For service protection and rights management, the service provider (e.g., JAM) will create a random service and random security port. Alternatively, the service port can be fixed or created from a human readable string, shown in Def. 4.

```
 1: // 1. Create random security and service ports
 2: var secret  = Port.unique(),
 3:     service = Port.unique();
 4: // 2. Put them together in a full rights capability
 5: var supercap = Capability(service,Private(0,0xff,secret));
 6: log(Capability.toString())
 7: // 3. Restrict capability
 8: var restrcap = Capability(service,
 9:   Private.restrict(supercap.cap_priv,0x20,secret));
10: var service2 = Port.ofString("12:34:56:78:90:12");
11: // 4. Check validatiy of client private part and rights
12: var cap = restrcap;
13: if (Private.rights_check(cap.cap_priv,0x20,secret)) {
14:   // grant service
15: }
```

*Def. 4. Service provider capability operations (code targets jamsh execution)*

### 3.6.3  Negotiation

Agents can negotiate specific resource constraints or the increase of the privilege level by using the `negotiate` function with a capability providing the respective rights to perform the operations, shown in Def. 5.

```
 1: // 1. A service port and a privare random port
 2: // must be created (randomly)
 3: var port = Port.unique()
 4: var rand = Port.unique();
 5: // 2. The port:rand key must be registered
 6: // on the JAM platform
 7: var security = {}
```

```
 8:  | security[Port.toString(port)]=Port.toString(rand);
 9:  | // 3. Create a restricted capability with some rights
10:  | var cap  = Capability(
11:  |  port,
12:  |  Private.encode(0,Rights.NEG_LIFE,rand))
13:  | // 4. Pass capability to agent
14:  └ create('foo',{cap:cap});
15:  ┌ // 5. In foo agent activity perform negotiation
16:  | // using this cap
17:  └ var success=negotiate('LIFE',10000,this.cap);
```

*Def. 5. Basic principles of resource and privilege negotiation using a capability (both JAM node and agent must rely on the same public service and secret random security ports)*

### 3.7  Agent Process Mobility and Migration

Mobility of agents require the transfer of the relevant state of the corresponding agent process and its program code. The agent process state consists of the control and the data state.

The control state of an agent is stored in a reserved agent body variable *next*, pointing to the next activity to be executed. Additionally, there is a hidden control state if the an agent activity processes a scheduling block (see Sec. 3.5). The data state of an AgentJS agent consists only of the body variables. There are no references to variables outside the agent process context (free variable references cannot be serialized).

Migration requires a snapshot of the agent process, in this case the agent itself, a serialization with code-to-text transformation, transportation of the text code to another logical or physical node, and a deserialisation with text-to-code conversion bound to a new sandbox-restricted environment, shown in Fig. 7.

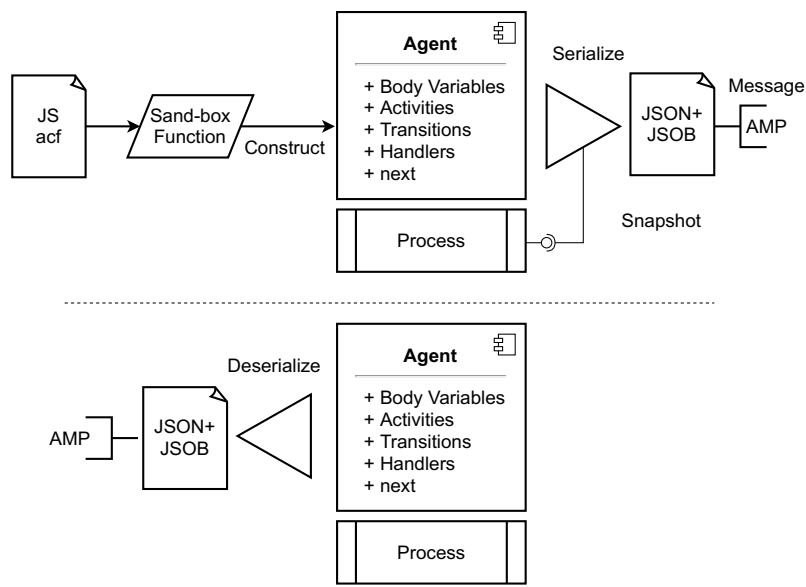

*Fig. 7. Agent process creation and process migration work flow*

The agent object is finally passed to the new node scheduler and can continue execution of the next activity or a next scheduling block entry. Text code sizes of medium complex agents (with respect to data and control space) are reasonable low about 10kB, simpler agents tend to 1kB, that can be significantly reduced by using LZ compression. One drawback of this method raises with pending scheduling blocks existing still in the snapshot. They must be entirely saved in the migrated snapshot, too, and back converted to code on the new node. Pending scheduling blocks contain function code and hence can increase the snapshot size significantly. Therefore, migration (using the `moveto` operation) requests should not be embedded in a scheduling block.

Agent migration is always performed via the Agent Management Port (AMP) with a negotiated virtual link connection between the sending and receiving node, discussed in the next section.

### 3.8 JAM Platform Connectivity

#### 3.8.1 JAM Networking

An agent can migrate to another JAM platform. JAM agents are mobile by transferring a snapshot of an agent process containing the entire data and control state including the program specifying the agent behaviour. Communication is a central part of agent interaction. Beside tuple space communication basically limited to the local node level, signal messages can be addressed to agents on remote platform, requiring a unified but loosely coupled platform connectivity.

JAM provides a broad range of connectivity options, shown in Figure 8, available on a broad range of host platforms. There is Peer-to-Peer (P2P) connectivity between neighbour nodes by using, e.g., serial links used in mesh-grid networks, and wide-area connectivity, i.e., Internet, by using generic HTTP, TCP. or UDP protocols via the AMP discussed in the next section and/or optionally by using the Distributed Organization System layer (DOS) and a broker server.

#### 3.8.2 Agent Management Port (AMP)

Using P2P connectivity, JAM nodes communicate via the *Agent Management Port* (AMP). AMP provides messaging between JAM nodes for agent migration, signal and tuple migration, collection of statistical data, and agent control. AMP messages can be transferred via any stream-like link. Additionally, an external monitor program (debugger) can connect to a JAM node via AMP. AMP is independent of any specific data communication protocol, and hence supports a wide range of communication technologies, not limited to IP communication, shown in Fig. 8.

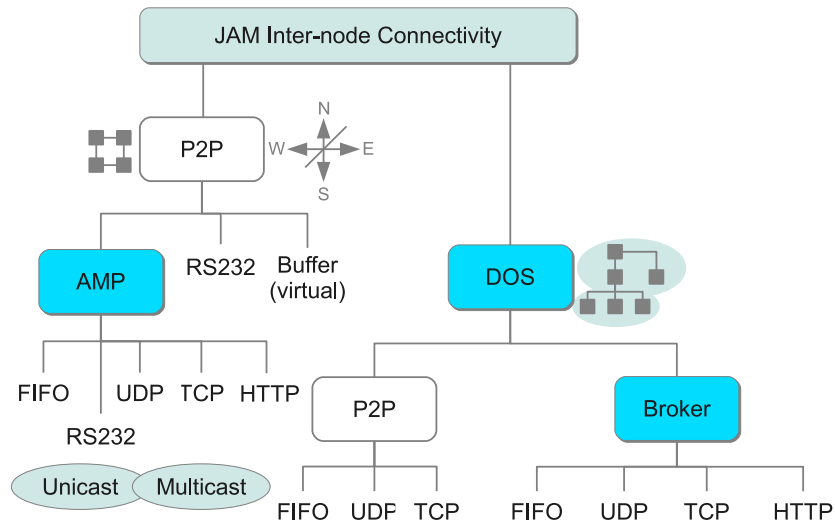

*Figure 8. JAM communication and node connectivity (P2P: Peer-to-Peer, AMP: Agent Management Port, DOS: Distributed Organization Layer)*

On the Internet, IP-based protocols are commonly used to provide AMP message passing between JAM nodes using UDP, TCP, or HTTP protocols. One common issue are private or virtual networks with Network Address Traversal (NAT). To establish UDP communication between NAT networks, an external public rendezvous broker providing UDP hole punching techniques can be used. In this case, JAM nodes register on the broker with their node name, and other JAM nodes can connect to the (IP hidden) JAM nodes by their node names. The broker supports domain services (partitioning of nodes in domains/groups, e.g., based on GPS data).

JAM nodes can be loosely linked via the Agent Monitor Protocol (AMP) creating a distributed platform network. AMP is used to connect JAM nodes, e.g., via the Internet by using generic HTTP communication encapsulating AMP messages. The AMP provides the

following message types:

| Message Type | Description |
|---|---|
| ACK | Acknowledgement reply |
| LINK | Unidirectional link negotiation request (pair-wise, can be protected by capabilities) |
| PING | Keep alive message between nodes to estimate the connectivity state |
| PONG | Reply to a *PING* message |
| UNLINK | Close a previously negotiated link |
| RPCHEAD | Header of a RPC message (support for connection-less protocols) |
| RPCDATA | Data chunk of a RPC message (support for connection-less protocols providing data fragmentation and ordering) |
| RPCHEADDATA | Header and data of a RPC message (only supported by connection-based protocols) |
| SCAN | Scan request for external services |
| INFO | Information request for external services |

*Tab. 1. AMP message types*

A platform *A* can be linked with a platform *B* by starting a link negotiation phase, which can be protected by a security capability. The initiator node sends a *LINK* message, that will be acknowledged by an *ACK* message and (if request is authorised) by another *LINK* message from *B* to *A*. To test the link status, *PING* and *PONG* messages are used (by both link partners). If a link is established (that may be unreliable and can vanish at any time), Remote Procedure Calls (RPC) can be executed by using the *RPCHEAD* and *RPCDA-TA* messages. A communication link between two nodes can be uni-directional (directed, e.g., $A \rightarrow B$). In this case, either two unidirectional links are set up by two nodes $A$ and $B$ independently, or the communication is used as is without acknowledgement. This operational state is still in accordance to the multi-agent model not relying on any reliable platform capabilities including communication. Although, connection links between nodes are established prior RPC messaging, the link does not rely on an underlying connection and state-based communication channel, e.g., like in the case of TCP connections. UDP never established a persistent communication channel, and HTTP do not require a keep-alive TCP connection. In contrast, the WebSocket transport layer always rely on a connection state and is therefore not primarily suitable for creating JAM networks.

The RPC messages encapsulate agent process snapshot migration, remote signal message delivery, and remote tuple space requests. Each AMP communication port can be a communication end-point or a router, too.

AMP enables the composition and connection of already distributed agent platform networks. Signal messages can be forwarded to agents on different remote nodes, but signal message propagation is limited to agent migration paths (traces), shown in Fig. 9. A signal from agent $A$ can only be delivered to agent $B$ if at any past time they joined the same node (this is always the case in parent-children groups). Agents on two different nodes are aware of the two different distinct nodes. Remote communication is not reliable, i.e., sending of signals or migration of agents can result in a total loss.

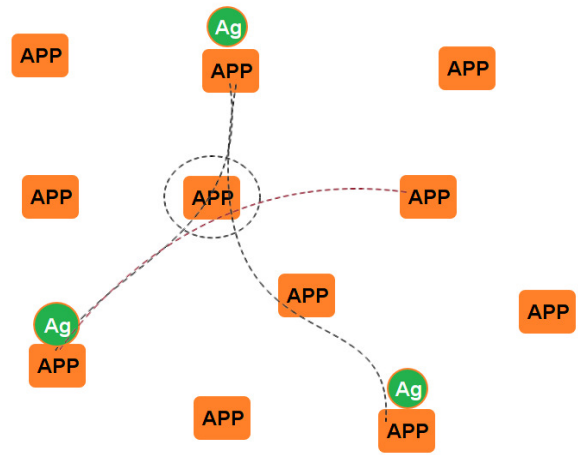

Fig. 9. Agent migration traces and crossing points (APP: JAM node). Signals can be delivered only along migration paths. Along a sub-path between two nodes there must be exist a negotiated AMP link.

### 3.9  Software Architecture and Workflow

The software architecture is composed of:

1. AIOS;

2. Modules (handling nodes, processes, communication, etc.);

3. Host platform API wrapper and utility function library;

4. Applications (shell, Web laboratory, simulator).

In Fig. 10, the basic software architecture with its components is shown. In addition, the basic operational workflow is sketched starting with the compilation of agent constructor functions from source text, agent process creation and scheduling, and the JAM world management.

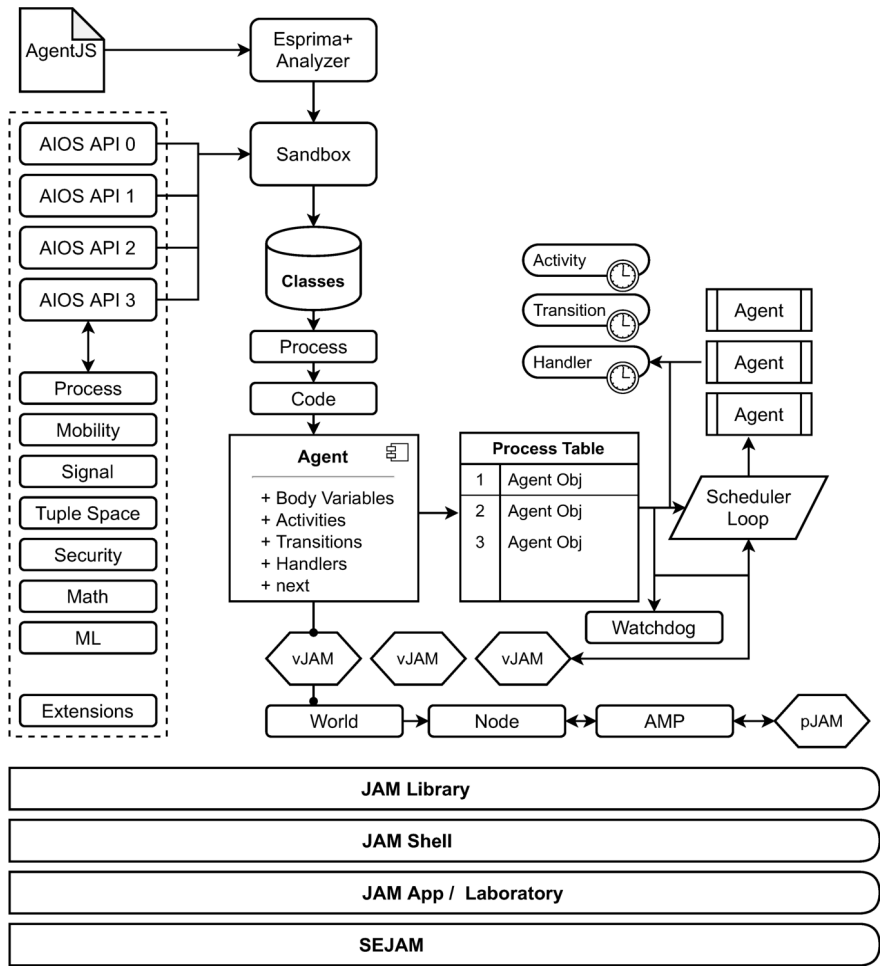

*Fig. 10. JAM components and software architecture with basic operational flow*

## 4. AgentJS API

AgentJS supports all ES 2015 JS compliant functions and statements. Due to the sandbox-restricted container encapsulation the `for(a in b)` operation is broken. Instead, AgentJS provides po-lymorphic iterator operations, introduced in the next Section. There is no support for objects with prototype function bindings (cannot be serialised efficiently). Objects in AgentJS are pure procedural, i.e., the agent or the AIOS has to provide separate functions to process objects (data records).

## 4.1 Computation

Computations can be performed in activities, transition and signal handlers. Computation primarily modifies agent data (agent body variables) accessed by the `this` object. The agent self-reference is available in all AIOS functions in callback functions passed to AIOS functions (at least on first level call context). Note that AgentJS only supports record data objects without prototype function bindings.

| Operation(s) | Description |
|---|---|
| `abs, add, angle, delta, distance, div, equal, max, min, neg, random, sum, zero` | Extended math functions |
| `concat, contains, copy, empty, filter, flatten, head, isin, iter, last, map, matrix, reduce, reverse, sort, tail, without, Vector` | Array and object functions |
| `int, object, string` | Type conversion |

*Tab. 2. Computational agent statements*

The iterator functions `concat. flatten`, and `iter` can be applied on array and record data structures, too. This is also the case for the functions `angle, contains, copy, delta, isin, neg,` and `zero`. Examples of polymorphic and mixed data type operations are shown in Ex. 4.

```
var a = [1,2,3,4];
var amin = min(a), amax = max(a);
var v = Vector(1,1);
if (zero(v)) v.x=v.y=1;
var b = [5,6,7,8]
var c = concat(a,b);
var d1 = { a:1, b:2 },
    d2 = { c:0 },
    d3 = concat(d1,d2),
    ds = 0;
iter(d3, function (e) { ds += e })
if (sum(d3)!=ds) kill(me())
if (isin(d3,2) && isin(a,2)) log('common element');
```

*Ex. 4. Examples of computational statements used in agent functions*

## 4.2 Agent Environment and Control

The agent is situated in its environment providing interaction and perception (data). The environment consists primarily of the agent and the host platform. The agent can request multiple sensors to get information about its platform, the location of the platform, process group relations, and about its state, summarised in Tab. 3. Finally, an agent has a privilege level providing restricted access to the JAM platform. The `privilege` operation returns the agent's current privilege level in the range [0,3] that can be changed by negotiation by using the `negotiate` operation, typically requiring a valid capability key with sufficient rights to negotiate resources with the platform.

| Operation(s) | Description |
|---|---|
| `info(cls)` | Generic sensor providing information about node, location, current agent processing constraints and limits. Node information can be requested by using *cls*='node'. This returns an object `{ id:string, position:{}, location:{ip:string ,gps:{} ,geo:{}}, type:string}` |
| `me` | Return the current agent identifier |
| `myClass` | Return the agent class (basically the name of the agent constructor function) |
| `myNode` | Return the current APP node name (identifier) |
| `myParent` | Return the agent parent (identifier) if this a forked child agent |
| `myPosition` | Return the current APP node position (if available) |
| `negotiate(what,val,cap)` | Change platform rights and restrictions |
| `privilege` | Return the current agent privilege level (0-3) |
| `time` | Current node time with adjusted world time lag (clock calibration and support for node processing pauses, e.g., in simulation mode) |

*Tab. 3. Environmental perception statements*

The following parameters (Tab. 4) can be negotiated by the agent (assuming appropriate rights given by the passed capability argument). Memory constraints cannot be controlled by JAM since there is no way on JS level to get information about the memory usage of specific objects, although, a normalised memory allocation factor assuming C-like memory allocation could be computed. But JAM is not aware of data allocations inside agent functions (under the VM GC control) and would require inspection of the data state after each function call, which would degrade JAM performance significantly. Only tuple space access, agent creation, and signalling can be controlled by JAM.

| Parameter | Rights | Description |
|-----------|--------|-------------|
| LIFETIME | NEG_LIFE | Change the agent lifetime (total accumulation) |
| SCHEDULE | NEG_SCHED | Change the maximal agent scheduling time |
| CPU | NEG_CPU | Change the maximal CPU time (sum of all schedule times) |
| TS | NEG_RES | Change the maximal number of tuple requests |
| AGENT | NEG_RES | Change the maximal number oif agents that can be created |
| LEVEL | NEG_LEVEL | Change the AIOS operational privilege level |

*Tab. 4. Platform and resource parameters that can be negotiated and teh required rights*

Some examples of environmental information and control operations are shown in the following Ex. 5.

```
me()
```
————

```
jireluci
```

```
myNode()
```
————

```
xazorodi
```

```
privilege()
─────────
1

info('node')
─────────
{ id: 'xazorodi',
  position: { x: 0, y: 0 },
  location:
   { ip: '31.16.196.110',
     gps: { lat: 53.0515, lon: 8.8896 },
     geo:
      { city: 'Bremen',
        country: 'Germany',
        countryCode: 'DE',
        region: 'HB',
        zip: '28309' } },
  type: 'shell' }

myPosition()
─────────
{ ip: '31.16.196.110',
  gps: { lat: 53.0515, lon: 8.8896 },
  geo:
   { city: 'Bremen',
     country: 'Germany',
     countryCode: 'DE',
     region: 'HB',
     zip: '28309' } }
```

*Ex. 5. Examples of the output from agent's environmental information functions*

### 4.3  Agent Management

Agents are either created by the platform and host application from a constructor function or by agents either from a constructor function or as a copy (child forking by a parent agent). Constructor function must be recompiled in a restricted context. To speed-up agent creation, agent constructor functions can be pre-compiled (and analysed) and added to the JAM platform.

### 4.3.1  Creation of Agents

Before an agent can be created either by the platform or by an agent the constructor function must be compiled (analysed) and stored in the platform class library dictionary. Agent forking can be done without an installed agent constructor function. The fork operations create an exact child copy of the parent agent with an optionally modified set of body variables.

| Operation(s) | Description |
|---|---|
| `create(class,parameter,level)` | Create an agent from a class constructor function (or alternatively it's string name of an already compiled agent class) with optional parameter settings. The AIOS level argument is optional any may not be higher than the level of the parent process |
| `fork(parameter)` | Create a copy of the calling agent with the same code and data state but (optionally) with different parameter settings (direct modification of body variables of forked agent) |

*Tab. 5. Creation of agents*

In the following Ex. 6 one agent is created programmatically by the JAM platform (using *jamsh*). Firstly, the agent constructor function is defined, finally compiled and added to the JAM platform. The agent starts in activity *init* remembering its identifier in its body variable *master*. A straight transition to the *replicate* activity forks a child agent with a modified set of body variables (*message*). The identifier of the newly create agent is store in the parent *child* variable. The child agent with move to another node via the *migrate* activity. If this variable is set, the parent agent will sleep after the *wait* activity was processed. The child agent (having still an empty *child* variable) will print out the message it has gotten from its parent and sleeps forever. After the parent is woken up it will kill its child agent. The kill signal is forwarded from the parent to the child node.

```
function ag (options) {
  this.child=null;
  this.master=null;
  this.message=null;
  this.act = {
    init : () =>) {
      this.master=me();
    },
    replicate : () => {
      this.child=fork({message:'I am John.'});
    },
    migrate : () => {
      var nodes = link(DIR.IP('%'));
      if (nodes.length) moveto(DIR.NODE(random(nodes)));
    },
    killing : () => {
      kill(this.child);
    },
    wait : () => {
      if (this.child) {
        sleep(1000);
      } else {
        log(this.message);
        sleep();
      }
    },
    end : () => {
      kill();
    }
  }

  this.trans = {
    init       : replicate,
    replicate  : () => { return this.child?wait:migrate },
    migrate    : wait,
    killing    : end,
```

```
   wait       : () => { return this.child?killing:wait },
 }

 this.next=init;

}
// Add agent class
compile(ag,{verbose:false});
// Create level 2 agent
create('ag',{ },2);
```

*Ex. 6. Examples of agent creation either by instantiating from a constructor function or by forking (cloning). Code is processed by the jamsh program.*

### 4.3.2  Modification of Agents

Agents can modify their behaviour ATG by adding or removing activities and transitions, shown in Tab. 6. Furthermore, activities and transitions can be modified (replacement). Transitions can be modified temporarily to create child agents with a different control flow. After an agent was forked the original transitions can be restored. An activity update or addition requires an unique activity name and a function. A transition rule update or addition requires the start activity and either a constant end activity name or a resolution function that can return any of the existing activities (or null if there is no active transition resulting in agent blocking). Some examples are shown in Ex. 7.

Transitions can be updated (replacing old transition rule using the *update* method) or extended with an additional transition selection function using the *add* function, see Ex. 7 for details. If the transition function references free variables (from current activity or handler function), the transition function can be parametrised with an additional data parameter, shown in the example "EXTEND" signal handler. For example, another agent can send a signal with a {act:string,fun:string} object argument containing a new activity name and the function implementation.

| Operation(s) | Description |
|---|---|
| `trans{.add .delete` `(start,end?,data?)` | Modification of transitions |
| `act{.add .delete .update}` `(name,function?)` | Modification of activities |

*Tab. 6. Modification of agents*

```
1:  function adapter() {
2:    this.v=0;
3:    this.act : {
4:      init : () => {
5:        // Uncondtional transition
6:        trans.add('more','wait');
7:      },
8:      replicate: () => {
9:        trans.update(replicate,migrate);
10:       this.child=fork({dir:DIR.EAST,MAXHOP:1});
11:       trans.update(replicate,wait);
12:     },
13:     percept : () => {
14:       trans.update('explore',function () {
15:         if (this.enoughinput<1) return 'goback';
16:         else return 'wait';
17:       });
18:
19:     },
20:     explore : () => {},
21:     goback: () => {},
22:     migrate : () => {},
23:     wait : () => {},
24:   }
25:   this.trans : {
26:     replicate : wait,
27:     choices : () => {
28:       return wait;
29:     },
30:   }
31:   this.on : {
```

```
32:     // Signal from other agent
33:     // data={act:string,fun:string}
34:     "EXTEND" : (data,from) => {
35:       data=JSON.parse(data); // JSON==JSON+
36:       act.add('choices-'+data.act,
37:             data.fun);
38:       // Extend choices transition rule
39:       // need to pass data to the conditional rule function
40:       trans.add('choices',(act) => {
41:         if (this.choice=='choices-'+act)
42:           return 'choices-'+act;
43:       },data.act);
44:     }
45:   }
46: }
```

*Ex. 7. Example usage of code morphing operations*

The ATG modification can be used to replicate lower functional agents by removing unnecessary activities and transitions temporarily. The original activity, transition, and handler functions can be saved in temporary variables. They can be either directly accessed by the agent `this` object or retrieved by the return value of the respective *delete* operation in case of activities and transitions. This is shown in the simple agent code in Ex. 8. The activity trace of the parent (root) agent is the run *a1* → *a2* → *a3*, whereas the forked child agent will execute the run *a1* → *a3*.

```
1: function foo (options) {
2:   this.child=null;
3:   this.act = {
4:     a1: () => {
5:       // var a2=this.act.a2
6:       var a2 = act.delete('a2');
7:       trans.update('a1','a3');
8:       this.child=fork({});
9:       act.add('a2',a2)
10:      trans.update('a1','a2');
11:    },
12:    a2: () => {},
13:    a3: () => {kill()}
```

```
14:    }
15:    this.trans = {
16:       a1:a2,
17:       a2:a3,
18:    }
19:    this.next = a1
20: }
```

*Ex. 8. Using temporary agent behaviour modification to create sub-class agents*

### 4.3.3 Control of Agents

Control of agents is primarily event-based and includes the termination of agents. Note that the termination performed by the *kill* operation only requires the agent identifier name, which should not be publicly visible. The agent control flow can be blocked explicitly either by waiting on a time event or by on an explicit wake-up performed either by an agents own signal handler or by another agent (on the same node), shown in Tab. 7.

| Operation(s) | Description |
| --- | --- |
| `kill` | Send the *kill* signal to an agent (without argument kills calling agent) |
| `sleep(tmo?)` | Suspend agent execution until a wake-up occurs (either explicitly or by a timeout in milliseconds). A suspended agent can still receive signals. |
| `wakeup(aid?)` | Wake up an agent, typically called from a signal handler (stop and go control). If no agent identifier is provided, this agent will be woken up (if suspend). |
| `timer.add(tmo,sig,arg,repeat)` | Add a timer to the agent process. After the time *tmo* in milliseconds elapsed, the signal *sig* will be raised with an optional argument *arg*. The agent must have a signal handler for this signal (*on* section). If the argument *repeat* is true, the timer is fired periodically. |
| `timer.delete(sig)` | Remove a timer from the agent process. |

*Tab. 7. Control of agents*

An example using `sleep` in conjunction with `wakeup` is shown in the following Ex. 9.

```
this.act : {
  init : function () {
    timer.add(100,'WATCHDOG');
  },
  delay : function () {
    sleep(1000)
  },
  await : function () {
    sleep();
  },
  process : function () {
    timer.delete('WATCHDOG');
  }
}
this.on : {
  'SIGNAL': function () {
    wakeup()
  },
  'WATCHDOG' : function (arg) {
    wakeup();
  }
}
```

*Ex. 9. Agent control with sleep and wake-up operations*

The *sleep* call in activity *delay* will suspend the agent processing (i.e., any further activity transition) until the given timeout occurs. The *sleep* call in the second activity *await* will suspend agent pro-cessing until a wake-up occurs, typically by receiving a signal from another agent, shown in the `SIGNAL` handler function. An additional background timer is started in activity *init*.

## 4.4 Agent Communication

Agents can communicate with each other by using undirected tuple spaces and directed signal messages, summarised in Fig. 11. Tuples can only be exchanged by agents processed by the same node. Signal can be send to agents processed on the same node or on separate nodes if there is an agent migration path between the source and destination node. Remote signal propagation can be used to access remote tuple spaces, too, basically to copy or transfer tuples between different nodes. Level 2/3 agents have access to the remote tuple space API.

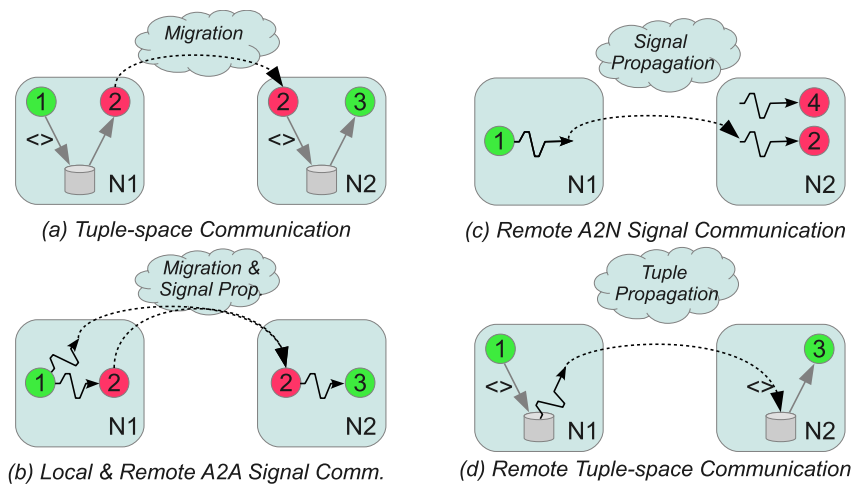

*Fig. 11. Basic agent-agent communication methods*

The following two sub-sections address the anonymous generative communication using tuple spaces and addressed communication using signal messages.

### 4.4.1 Generative Tuple-space Communication

A tuple is a data container that binds multiple values. Formally, tuple elements cannot be accessed directly, in contrast to arrays. But AgentJS implements tuples with arrays. In contrast to arrays, tuples are poly-typed, i.e., each element can have a different data type. Arrays in JS are poly-typed, too, and be created on the fly (as tuples). A tuple is characterised by its arity and its type signature (of all its elements).

Tuples are stored in a tuple space, basically a non-indexed data base, i.e., tuples canot be accessed directly in the data base. Instead patterns are used to access a subset of matching tuples.

There are basically two basic operations on tuple spaces:

**out(t)**

A new tuple *t* is stored (written) in the tuple space data base. The tuple is fully decoupled from its source agent (there is nor information about the source). A tuple exists for a limited time. There can be multiple equally tuples (multiple copies).

**inp(p)**

A tuple is removed from the tuple space data base based on a tuple pattern *p*. The input operation is synchronous and if there is no matching tuple the calling agent process will be blocked (suspended) until a matching tuple is stored in the data base (by another agent or by the host application). The input operation is atomic. At most one tuple is removed.

In addition to the input operation, there is a read operation that do not removes a matching tuple. Instead a copy is returned. The basic tuple space API is shown in Tab. 8. Blocking operations need a callback handler function that is called if there is a matching tuple found or a time-out occurs. A tuple is an array consisting of actual values (number, string, data record, data array) or a formal parame-ter represented by a null value. Formal parameters are used in patterns, only.

| Operation(s) | Description |
|---|---|
| `alt(p [],function),`<br>`alt.try(tmo,[[]],function)` | Scan for mutiple patterns. The first tuple matching one of the patterns is removed from the tuple space and passed to the handler function. |
| `exists(p)` or `test(p)` | Check for the existence of a tuple |
| `inp(p,cb),`<br>`inp.try(tmo,[],cb)` | Remove a tuple from the data base based on pattern matching. The try-version waits a limited time for a matching tuple (or returns immediately with a null value if there is no tuple). |
| `mark(p,tmo)` | Store a temporary tuple in the data base with defined time-out |
| `out(t)` | Store a tuple in the data base |
| `rd(p,cb)`, `rd.try(tmo,[],cb)` | Read a tuple (copy) based on pattern matching. The try-version waits a limited time for a matching tuple (or returns immediately with a null value if there is no tuple). |
| `rm(p,all)` | Remove one or all tuples matching the given pattern. |
| `ts(p,callbackOrTimeout)` | Update a tuple by atomic test and set operation (in-place modification). If teh second argument is a function, the tuple is replaced by the returned tuple, if it is a number than the lifetime of a tuple is set or extended by this time value. |

*Tab. 8. Local tuple space operations (p : pattern, t : tuple)*

The `alt` operation allows the listening on multiple tuple patterns. The first tuple that matches a pattern activates the handler function.

There is one tuple handler function that is used for all tuple requests. The *alt* operation behaves like the *inp* operator by removing the matching tuple (atomic operation). The test operation can be used to test for a tuple based on a pattern. The test operation returns true if there is actually a matching tuple. After a test operation, a read or input operation can be performed. Since the agents of one physical node are scheduled sequentially, there is no consistency issue by violating atomicity between test and input. Remote tuple spaces can be accessed if the agent has the appropriate rights (privilege level) and if there is a link between its node and the remote node, shown in Tab. 9.

| Operation(s) | Description |
|---|---|
| `collect(to,p)` | Migrate all matching tuples form this source TS to the remote destination TS. |
| `copyto(to,p)` | Copy all matching tuples form this source TS to the remote destination TS. |
| `store(to,t)` | Store a tuple *t* on node *to* (node identifier). |

Tab. 9. Remote tuple space operations requiring a link between source and destination nodes.

Examples of tuple space operations are shown in the following Ex. 10 code snippet. The operations are used in agent activities. Blocking operations may not be used in transition and signal handler functions. Operations with timeout can return a null value, i.e., callback functions have to check for valid tuples.

```
alt.try(500,[
  ['SENSOR',null],
  ['TIME',null],
  ['NODE',100,null]
```

```
],(t) => {
  if (!t) log('timeout');
  if (lof
})
out(['SENSOR','temp',5000])
mark(['IAMHERE',myid(),mynode()],1000 /*ms*/)
rd(['SENSOR',null],(t) => {
  if (t) this.sensor=t[1];
})
if (test(['SENSOR',100])) { .. }
ts(['SENSOR',100], (t) => {
  t[1]++;
  return t
})
```

*Ex. 10. Examples of agent tuple operations*

### 4.4.2  Signals

In contrast to generative and anonymous tuple communication, signal messages are always sent to a destination with a known address (commonly the agent identifier). Signals are lightweight messages that are sent between two agents *A* and *B*. The agent can be on different nodes, but at some time in the past they had to be processed by the same node (the crosspoint node). Typically, the agents are related by first-order parent-child trees (e.g., *B* is a child of *A* created by forking or cold instantiation from a constructor function). If an agent migrates to another node, the path is remembered for some time (with a node-specific time-out). The delivery of signal messages relies on the migration tables if there is no local destination agent and the message is forwarded to next node until no further path elements can be resolved or a maximal hop-count is reached. An example of signalling between parent-child agents is shown in Ex. 11. The script code must be executed by the JAM shell program.

| Operation(s) | Description |
|---|---|
| `send(to,sig,arg)` | Send a signal message of type *sig* to agent *to* with an optional argument *arg*. |
| `broadcast(ac,range,sig,arg)` | Send a signal message of type *sig* to agents of class *ac* in the given range hops with an optional argument *arg*. |
| `on."SIGNAL" : function (arg,from)` | Signals can only received by installed agent signal handlers (in the *on* section). The signal name is the handler name attribute. |

*Tab. 10. Signal operations*

In the following Ex. 11, a simple ping-pong signal messaging example between a parent-child agent pair is shown. Two agents are created, passing the identifier name of the first to the second agent, enabling the sending of addressed signal.

```
1: function ac(config) {
2:   this.parent = config.parent;
3:   this.act = {
4:     init: () => {
5:       log('START',this.parent);
6:       if (this.parent)
7:         send(this.parent,'PING',random(0,100));
8:     },
9:     wait : () => { sleep(5000) },
10:    end : () => { log('END'); kill() }
11:  }
12:  this.trans = {
13:    init : wait,
14:    wait : end,
15:  }
16:  this.on = {
17:    'PING' : (arg,from) => {
18:      log('PING',arg,from);
19:      send(from,'PONG',random(0,100));
```

```
20:      },
21:      'PONG' : (arg,from) => {
22:        log('PONG',arg,from);
23:      },
24:    }
25:    this.next = 'init';
26: }
27: compile(ac,0,true);
28: var ag1 = create('ac',{});
29: var ag2 = create('ac',{parent:ag1});
30: start()
```

*Ex. 11. A simple ping-pong signal messaging example between a parent-child agent pair.*

## 4.5 Agent Mobility

An agent can migrate from its current to another (remote) node. This operation requires an established virtual link between the source and destination node. A node link can be tested by using the `link` operation. The `link` operation can be used to get all current links, too. In IP networks, the `link(DIR.IP("%"))` call returns the node identifier names, and `link(DIR.IP("*"))` returns the URL addresses of the currently linked nodes.

| Operation(s) | Description |
|---|---|
| `moveto(dir)` | Send this agent to new destination node specified by *dir*. |
| `link(dir)` | Test a link to a node or returns current link information. |
| *dir* ∈ `DIR` | Link destination direction object |

*Tab. 11. Agent mobility operations; link directions are defined below.*

```
type DIR = {
  NORTH, SOUTH,
```

```
WEST,   EAST,
LEFT,   RIGHT,
UP,     DOWN,
ORIGIN,
NW, NE, SW, SE,
BACKWARD,
FORWARD,
OPPOSITE,
// Assuming:  z-> x      N
//            |         W+E   U(+z)/D(-z)
//            v y        S
DELTA: function (addr) → {tag:"DIR.DELTA",delta:addr},
// Only for link? operation
RANGE: function (r) → {tag:"DIR.RANGE",radius:r},
// Address a node (identifier name) directly
NODE: function (node) → {tag:"DIR.NODE",node:node},
// Patterns: %:Node names; *:URLs
IP:function (addr) → {tag:"DIR.IP",ip:addr},
// Path can contain filters,
// e.g. range /distance[0-5], /distance[5], ..
// or a pattern for a set of destinations, e.g., /node*
// or a hopping array [dest1,dest2,..]
// type of path = string | string array
PATH:function (path) → {tag:"DIR.PATH",path:path},
CAP:function (cap) → {tag:"DIR.CAP",cap:cap}
}
```

*Def. 6. Possible link directions used by the connect, move-to, and link operations. The first set is used only in mesh-like networks (primarily in virtual node networks like in the simulator)*

A simple moving agent is shown in Ex. 12. The currently linked nodes are returned by the `link(DIR.IP('%'))`. A node is randomly chosen. The agent migrates to the new node in the activity *move*, and back to its root node in the activity *back*.

```
function ag() {
  this.goto=null;
  this.root=null;
  this.act = {
    check: () => {
      this.root=myNode();
```

```
    var links=link(DIR.IP('%'));
    if (links && links.length)
      this.goto=random(links);
    else sleep(500);
  },
  move: () => {
    moveto(DIR.NODE(this.goto))
  },
  back: () => {
    moveto(DIR.NODE(this.root))
  },
  end: () => {
    log('Terminating');
    kill();
  },
  }
  this.trans = {
    check:() => {
     return this.goto?'move':'check' },
    move:'back',
    back:'end' }
  this.next = 'check';
}
```

*Ex. 12. An agent visiting a neighbour node and returning to its root node.*

## 4.6  Machine Learning

The fusion of agent and machine learning models creates learning agents. A ML module part of JAM provides a broad range of machine learning algorithms including reinforcement learning. The ML module provides the algorithms. Most (but not all) models used by the algorithms are pure non-cyclic data objects that are mobile, i.e., the models can be carried by migrating agents and further processed on another node.

An agent (with at least privilege level 1) can access the ML module by the `ml` root object. The type signature of the ML module is shown in Def. 7. The ML module consists of data analysis functions (`stats` object) and a set of generic ML operations, mainly *learn* creating a ML model from data performing training immedi-ately (e.g., in the case of decision trees), *train* to apply training to an already created model (e.g., ANN), and *predict* to apply a trained model om new data. Some models can be updated (e.g., Reinforcement Learner) by the *update* operation. There is a large set of ML algorithms covered by the generic operations, which can be selected by the `ml.ML` entries.

```
type ml = { action: [Function: action],
  classify: [Function: classify],
  compact: [Function: compact],
  depth: [Function: depth],
  evaluate: [Function: evaluate],
  info: [Function: info],
  learn: [Function: learn],
  learner: [Function: learn],
  noise: [Function: noise],
  preprocess: [Function: preprocess],
  print: [Function: print],
  similarity: [Function: similarity],
  stats:
   { analyze: [Function: analyze],
     entropy: [Function: entropy],
     entropyN: [Function: entropyN],
     entropyEps: [Function: entropyEps],
     entropyTEps: [Function: entropyTEps],
     entropyT: [Function: entropyT],
     features: [Function: features],
     gainEps: [Function: gainEps],
     maxGainEps: [Function: maxGainEps],
     mostCommon: [Function: mostCommon],
     partition: [Function: partition],
     partitionEps: [Function: partitionEps],
     partitionUniqueEps: [Function: partitionUniqueEps],
     splitEps: [Function: splitEps],
     unique: [Function: unique],
     uniqueEps: [Function: uniqueEps],
     utils:
      { best: [Function: best],
        bestNormalize: [Function: bestNormalize],
        column: [Function: pluck],
        log2: [Function: log2],
```

```
        prob: [Function: prob],
        relax: [Function: relax],
        select: [Function: select],
        selectEps: [Function: selectEps],
        sort: [Function: sort],
        stat: [Object],
        without: [Function: without],
        wrap: [Function: wrap] } },
test: [Function: test],
update: [Function: update],
// supported model/algorithm types
ML:
  { ANN: 'ann',          C45: 'c45',
    CNN: 'cnn',          ICE: 'ice',
    DTI: 'dti',          ID3: 'id3',
    KMN: 'kmeans',       KNN: 'knn',
    KNN2: 'knn2',        MLP: 'mlp',
    RF: 'rf',            RL: 'rl',
    SLP: 'slp',          SVM: 'svm',
    TXT: 'txt',          EUCL: 'euclidean',
    PEAR: 'pearson',     DPAgent: 'DPAgent',
    TDAgent: 'TDAgent', DQNAgent: 'DQNAgent' },
predict: [Function: classify],
train: [Function: learn],
best: [Function: best] }
```

*Def. 7. The type signature of the AgentJS Machine Learning module*

Some selected models and algorithms are presented in the next sub-sections.

### 4.6.1  Decision Tree

There are different decision tree learners, mainly C45, ID3, and a confidence interval decision tree model DTI based on interval arithmetic. The interval arithmetic is used to process input variables with a noise interval [x-εx+ε]. Values with overlapping intervals are considered as equal and do not contribute to an information gain. The noise (variable variance or standard deviation) values must be provided by the user. Either one value is provided for all input variables or one value for each input variables (array of ε values).

```
xy=[{a:..,b:,..,..,y:..},{...},...]
model = ml.learn({
 algorithm:ml.ML.C45, // ID3
 data:xy,
 target:'y',
 features:['a','b',..]
})
result = ml.classify(model,xy[0])
```

*Def. 8. Basic C45/ID3 template (input variables can be number, string, or Boolean values, output variable is always a string type)*

```
x=[[...],[...],..]
y=['a','b',..]
model = ml.learn({
  algorithm:ml.ML.DTI,
  x:x,
  y:y,
  eps:number // or [e1,e2,...]
});
```

*Def. 9. Basic DTI template (input variables may only be numbers, output variable is always a string type)*

### 4.6.2 Multi-layer Perceptron

The MLP model is portable (mobile) and implements a simple neural network with hidden layers. The model can be trained on creation using the `ml.learner` function at once or iteratively using the *ml.train* function. Training of MLP models can be time consuming and should be split into multiple activity runs (otherwise the activity processing can be terminated by the scheduler). The return of the training function delivers the current *loss* value and the time consumed by the training run. More information can be found in [23].

```
x=[[...],[...],..]
xo=[{...},{...},..]
y=['a','b',..]
model = ml.learner({
        x:x, // alternative: xo,
        y:y,
        features:['x1','x2',...],
        labels:['a','b',...],
        hidden_layers : [4,4,5],
        algorithm:ml.ML.MLP,
        normalize:true,
        bipolar:true,
        verbose:1
      });
result = ml.train(model,{
   epochs:1000,
}) // -> { time, epochs, loss }
result = ml.classify(model,x[0])
```

*Def. 10. Basic MLP template*

### 4.6.3 Reinforcement Learner

The Reinforcement Learner (RL) is closely related to the concept of the adaptive behaviour of agents. The RL trainer needs an environment with helper functions. The state is a numerical value as well as the reward. More information can be found in [24]. The action itself is not visible to the RL and is considered symbolic. The model is portable (pure data).

```
actions = ['a','b',..]
env = {
```

```
  // return allowed actions in current state
  allowedActions : (state) => {
  },
  // state transition functioin
  nextState : (state,action) => {
    return state
  },
  getNumStates : () => { return number},
  getMaxNumActions : () => { return number},
  // compute the reward of the state transitions and action
  reward : (state,action,nextstate) => {
    return number
  },
}
model = ml.learn({
  algorithm   : ml.ML.RL,
  kind        : ml.ML.TDAgent,
  actions     : actions,
  // specs
  // function learning rate
  alpha       : 0.1,
  // learning rate for smooth policy update
  beta        : 0.2,
  // initial epsilon for epsilon-greedy policy, [0, 1)
  epsilon     : 0.2,
  // discount factor, [0, 1)
  gamma       : 0.5,
  // eligibility trace decay, [0,1). 0 = no eligibility traces
  lambda      : 0,
  // number of planning steps per iteration. 0 = no planning
  planN       : 5,
  replacing_traces : true,
  smooth_policy_update : false,
  update : 'qlearn',  // 'qlearn' or 'sarsa'
  environment : env
})
while () {
  action = ml.action(model,state);
  next = env.nextState(state,action);
  reward = env.reward(next)
  ml.update(model,reward)
  state = next
}
```

*Def. 11. Basic RL template*

## 5. JAM for Users and Programmers

### 5.1 JAM Library

The library version of *JAM* contains a full agent processing VM with the *AIOS* as the central component enabling embedding of *JAM* in any host application including Web HTML pages. A public API is used to create and control *JAM* node instances. Each instance is an object that provides a set of API methods to control the JAM instance and to connect the instance to other nodes. There are two slightly different versions of the *jamlib* targeting:

1. Server side and terminal version supporting node.js [27], jerryscript [28], and quickjs [29];

2. Client side version supporting any Browser that provides JS processing and Cordova-based applications for Smartphones.

A JAM object instance is created by calling the `Jam` constructor function, shown in the code template in Def. 12. A selection of JAM library object methods (instantiated by the *Jam* constructor function) is shown in Tables 12 and 13.

```
var JAM = require('jamlib.js');
// Create a JAM instance with a virtual world
// containing one initial virtual node
var jam    = JAM.Jam({});
jam.init();
// Start the JAM scheduler loop
jam.start();
// Define an agent constructor function
function ac1(p) { this.x=p; this.act={}; this.trans={}; .. }
// Compile the agent constructor
JAM.compileClass(ac1,{verbose:1});
// Instantiate agents from class constructor
// with specific AIOS level
var a1 = JAM.createAgent('ac1',[args],1);
```

```
var a2 = JAM.createAgent('ac1',[args],1);
```

*Def. 12. JAM node instantiation and agent creation using the JAM library*

The *compileClass* function creates and installs the sandbox-restricted agent constructor functions for each privilege level with a different operational AIOS function set. The *createAgent* function instantiates an agent from the already sandbox-restricted agent class constructors with a specific parameter set and a specific AIOS level (here 1).

The AIOS can be easily extended using the *extend* method, e.g., adding natural language processing or logic solver APIs.

| Method | Parameter | Description |
|---|---|---|
| addClass | *name, constructor, env* | Add agent class to the JAM world and create sandbox constructors. |
| addNode | *nodeDesc*: {x,y,id?} | Add a new node to the world. Assumption: 2D mesh-grid network with (x,y) coordinates. The root node has position {x=0,y=0} |
| analyze | *ac, options* | Analyze agent class template in text or object form |
| compileClass | *name, constructor, options* | Analyse and compile an agent class constructor function and add it to the world class library |
| connectNodes | *nodes*: {x,y} [] | Connect logical nodes (virtual link) |
| connected | *dir, nodeid*? | Check connection status of a link |
| `createAgent,`<br>`createAgentOn` | (*nodeid*), *ac*, *args*, *level*, *className*, *parent*? | Create and start an agent from class ac with arguments |
| `createPort` | *dir*, *options*, *nodeid* | Create a physical communication port |
| `disconnect` | *dir*, *nodeid* | Disconnect remote endpoint |
| `extend` | *level*, *name*, *funcOrObj*, *argn* | Extend the AIOS agent API |
| `info` | *kind*, *id* | Get info about nodes, agents, and the host platform |
| `kill` | *id*, *nodeid* | Kill agent with specified id ("*": kill all agents on node or current node) |
| `locate` | *nodeid*, *callback*, *options* | Try to locate this node (based on network connectivity) and get geospatial position |
| `inp, mark, out, rd, ts` | - | Tuple space operation applied to current node |
| `saveSnapshot,`<br>`saveSnapshotOn` | *agentid*, *nodeid*, .. | Take an agent process snapshot executed currently on given node (returned as string or saved in file) |
| `setCurrentNode` | *nodeid* | Set current v-JAM node |
| `signal` | *to*, *sig*, *arg*, *broadcast* | Send a signal to a specific agent or broadcast |

*Tab. 12. Selection of JAM library operational methods*

| Method | Parameter | Description |
|--------|-----------|-------------|
| `init` | *callback* | Create and initialize node(s)/world, add optional TS provider/consumer, create physical network connections |
| `schedule` | - | Force a scheduler run immediately normally executed by the jam service loop. Required if there were external agent management, e.g., by sending signals |
| `start0`, `start`, `stop` | *callback* | Start or stop JAM (0: without scheduler loop for single stepping) |
| `step` | *steps*, *callback* | Stepping the scheduler loop |

*Tab. 13. JAM library control methods*

The JAM run-time supports two scheduling modes: Free running and single stepping. In free running mode, the scheduler determines the next schedule loop execution time based on pending events and ready agents (with a pending transition). In single step mode, only a fixed number of scheduler loop runs will be processed. This feature is used for debugging and simulation. The scheduler can only calculate the next loop execution time on events originating from agents (e.g., by calling the *sleep* operation). Externally introduced events require a forced scheduler run (by using the *schedule* operation).

## 5.2  JAM Shell

The JAM shell (`jamsh`) is a command line API for JAM basically providing a REPL and script wrapper to the JAM library. It provides a typical script-based programming interface to the underlying agent machine, easing creating and controlling agents, composing JAM networks and connecting JAM nodes, and so on. JAM scripts can be either loaded and processed directly from a file and/or by typing commands in the REPL shell. Agent class constructors can be loaded from files, compiled from functions or any text representation. There is simplified and restricted Web version of the JAM shell, too, basically with restrictions due to connectivity (HTTP only) and missing watchdog protection.

### 5.2.1 Shell Commands

Shell commands either provided by the shell command line or executed in a script file are summarised in Tab. 14. Most functions and procedures are directly related to the AIOS and *jamlib* API.

| Command(s) | Description |
|---|---|
| `add(x,y)` | Add a new logical (virtual) node |
| `agent(id,proc?)` | Returns the agent object (or process) |
| `agents()` | Get all agents (id list) of current node |
| `args` | Script arguments |
| `ask(question,choices)` | Ask a question and read answer. Available only in script mode. |
| `array, assign, contains, concat, copy, empty, filter, last, merge, neg, object, pluck, random, select, without` | Array, matrix, object, and math functions |
| `Capability, Port, Private` | Create a security capability object, port, and private field |
| `clock(fmt)` | Return system time (ms or hh:mm:ss format) |
| `cluster(desc)` | Create a worker process cluster |
| `config(options), configs` | Configure JAM. Options: print, printAgent, TSTMO, .., and return configuration of JAM AIOS |
| `connect({x,y},{x,y}),`<br>`connect(to),`<br>`disconnect({x,y},{x,y}),`<br>`disconnect(to),`<br>`connected(to)` | Connect two logical nodes (DIR.NORTH...), connect to a physical node, disconnect, and check a connection to specified destination. |
| `compile(f,options),`<br>`open(file)` | Compile an agent class constructor function; open and compile agent class file; |
| `create(ac, args,`<br>`level,node), kill(id)` | Create an agent from class @ac with given arguments @args and @level optionally on @node (return agent id); kill an agent |
| `extend(level,name,argn)` | Extend the AIOS for one or more privilege levels |
| `http, https` | HTTP and HTTPS client and server objects supporting get, GET (JSON), put, PUT (JSON), and server operations. |
| `info(what,id)` | Return node information object ('node', 'version', 'host') |
| `inp(p,all), mark(t,tmo),`<br>`out(t), provider(cb),`<br>`rd(p,all), rm(p,all),`<br>`test(p)` | Read and remove (a) tuple(s) from the tuple space (non-blocking); store a timeout-limited tuple; store a tuple in the current node TS; installation of a tuple provider and consumer handler; read a tuple from the TS; remove tuples; test for tuple(s); |
| `later(tmo,cb)` | Execute a function later. If the function returns a true value, the next timer is started. |
| `node(id), nodes, name, world` | Get or set current v-node; return all v-node ids; return name of current node; return current world object |

| `on(ev,handler)` | Install an event handler. Events: "agent+","agent-","signal+","signal","link+","link-","exit" |
|---|---|
| `port(dir, options:{proto,secure}, node)` | Create a new physical communication port |
| `setlog(flag,val)` | Enable or disable logging features |
| `signal(t,sig,arg)` | Send a signal to specified agent |
| `sleep(milli)` | Suspend entire shell for seconds |
| `start(), stop()` | Start and stop JAM scheduler (and other services) |
| `uniqid(options)` | Return a random name |
| `utime()` | Return high resolution system time in nanoseconds |

*Tab. 14. Common JAM shell commnds*

### 5.2.2 Ports and Links

A link between two (physical) nodes requires the set-up of a communication port using the `port` operation. Typically, IP networks are used to establish JAM node links. At least one node needs a static IP port. A link is established by using the `connect` operation, shown in Def. 13.

```
// Node A
var sport = Port.ofString('12:34:56:78')
port(DIR.IP('http://localhost:4567'))
port(DIR.IP('http://localhost:5567?secure=12:34:56:78'))
port(DIR.IP('http://localhost:6567'),{secure:sport})
// Node B
port(DIR.IP('http://localhost:*'))
connect(DIR.IP('http://localhost:4567'));
```

*Def. 13. Creation of IP ports and establishment of links between nodes. Ports can be protected by capability ports or by any string secret key*

Instead using HTTP for AMP communication, UDP and TCP can be used as well. The protocol can be directly set by the direction IP URL, e.g., `port(DIR.IP('udp://localhost:4567'))`. Note that Web browser only support HTTP communication. A JAM shell can act as an agent relay node just by creating IP ports publicly visible. Any JAM node with Internet connectivity can connect to this node via AMP the *connect* operation. JAM nodes can be connected by (level-3) agents, too, using the *connectTo* operation.

### 5.2.3 Agent Classes

An agent constructor function can be compiled from am embedded function definition, a function definition loaded from a source file, or from a text string containing the code. A compilation is performed by the `compile(ac, name?, options?)` function adding the sandbox-restricted constructor functions for all AIOS privilege levels to an internal library. Constructor functions can also be loaded directly from a file by using the `open(file,verbose?)` function. An agent class may only contain one named constructor function. The *compile* function uses a modified esprima JS analyser reflecting specific AgentJS features and that is applied to the agent constructor function. Analysis results can be displayed by setting the options attribute *verbose*.

```
function ac (p) {
  this.v=ε;
  this.act={..};
  this.trans={..};
  this.on={..};
  this.next='..' }
compile(ac,{verbose:true});
// Alternatevily read acf from file
// myclass.js contains: function myclass(p) {..}
open('myclass.js',true);
```

*Ex. 13. Compiling of an agent constructor function*

### 5.2.4 Physical Clusters and Virtual Worlds

The JAM shell can be used to easily create worlds consisting of virtual or physical JAM nodes connected by virtual circuit links or IP-AMP links. Physical JAM nodes are grouped in clusters and can be created from a root JAM shell instance.

Each JAM node is a world container containing at least one virtual node with its own process tables and tuple space data bases. Additional virtual nodes can be added to the JAM world. Virtual nodes of one world can be connected via virtual channel links with a two-dimensional mesh-grid network structure. Communication (agent migration and signal propagation) is supported via North, South, West, and East communication ports. The virtual node network is primarily used for simulation (see Sec. 7) and virtualisation of networks. The *jamsh* program provides direct support for virtual JAM networks, shown in Def. 14. Each node is identified by its logical position in the 2D world $(x, y)$. The initial root node is at position (0,0). Commonly, the world and root name are the same (but, initially, the world name is always in upper case letters).

```
// Create virtual 2x2 node network
var node2 = add({x:1,y:0}),
    node3 = add({x:1,y:1}),
    node4 = add({x:0,y:1});
// unidirectional links!
connect({x:0,y:0},{x:1,y:0})
connect({x:1,y:0},{x:1,y:1})
connect({x:1,y:1},{x:1,y:0})
connect({x:1,y:0},{x:0,y:0})
function foo() {
  this.act = {
    foo : () => {
      // migrate in virtual network
      var dirs = [DIR.NORTH,DIR.SOUTH,
                  DIR.EAST,DIR.WEST],
          goto=null;
```

```
      iter(dirs, (dir) => {
        if (!this.goto && link(dir))
          this.goto=dir;
      })
      if (goto) moveto(goto);
    }
  }
}
```

*Def. 14. Creation of virtual node networks processed by the same physical node in the JAM shell*

In contrast to virtual node networks, physical clusters consisting of independent JAM node instances connected via IP-based communication channels (using AMP). The JAM shell support creation of JAM clusters directly, shown in Def. 15. again, nodes are arranged in a two-dimensional mesh-grid. A cluster creates a set of IP-AMP ports. There are external and internal ports. Internals ports are used to connect the nodes directly via directional links, i.e., using North, South, West, and East ports. External (not directional) ports support connection from outside, e.g., using HTTP from Web browsers. Note that the root JAM node executing the shell script is not the first node in the cluster! Therefore, the root node must connect to one of the cluster nodes via an external port.

A physical cluster is created by the JAM shell by using the `cluster` operation defining the grid network (rows and columns), the AMP IP ports and protocols, base ports for the AMP IP ports for each link class (p2p, multi), a poll interval to check for cluster node state, and an initial shell command to be executed by each cluster worker (*todo*), shown in Def. 15.

```
// Create physical 2x2 node cluster network
var workers = cluster({
  connect:true,
  rows:2,
  cols:2,
  port0:11001, // First internal IP port
  port1:10001, // First external IP port
  portn:100,   // Increment
```

```
  proto:['http','udp'],
  poll:number, // worker poll interval in ms
  todo:'config({log:{node:true}});start()',
  verbose:0,
});
workers.start()
// Connect to first cluster node
var p1=port(DIR.IP('udp://localhost:12000'));
later(500,() => {
  connect(DIR.IP('udp://localhost:10002'))
});
function foo () {
  this.act = {
    tofirst : () => {
      // find connected cluster node
      var connected=link(DIR.IP('%')),
          goto=null;
      if (connected && connected.length)
        goto=DIR.NODE(connected[0]);
      if (goto) moveto(goto);
    },
    toright : () =>  {
      // migrate in cluster
      moveto(DIR.EAST)
    }
}
// Create an agent that migrate from this node
// to the first cluster node
create(...)
```

*Def. 15. Operational template to create physical JAM cluster networks using the JAM shell*

Agents can migrate between cluster nodes by using the `DIR.NORTH` ... direction flags. Each cluster node creates public IP ports (starting with the IP port number *port0*, e.g., 11001) and private directed links (North, South, West, East), starting with IP port *port1*, e.g., 10001. The cluster is created and controlled by an initial JAM shell instance (not being part of the cluster network). Agents can be created on this controller instance. The agents can migrate to a node of the cluster if the controller instance has a link to one of the public ports of a cluster node. This is shown in the following Ex. 14.

```javascript
function circulator(maxhop) {
  this.maxhop=maxhop;
  this.hops=0;
  this.last=null;
  this.goto=null;
  this.act = {
    start: () => { log('Start on '+myNode()); sleep(1000); },
    move : () => {
      var next;
      // Find directed links between cluster nodes
      if (link(DIR.EAST)) next=DIR.EAST;
      else if (link(DIR.SOUTH)) next=DIR.SOUTH;
      else if (link(DIR.WEST)) next=DIR.WEST;
      else if (link(DIR.NORTH)) next=DIR.NORTH;
      // still on controller node?
      else if (link(DIR.IP('%')))
        next = DIR.NODE(link(DIR.IP('%'))[0]);
      log(next)
      this.goto=next;
      if (next) moveto(next);
    },
    hello : () => { this.hops++; log('I am on '+myNode()) },
    end   : () => { log('END'); kill() },
  }
  this.trans = {
    start : move,
    move  : () => { return this.goto?hello:end },
    hello : () => { return this.hops<this.maxhop?move:end }
  }
  this.next='start';
}
compile(circulator,{verbose:false});
later(1000, () => {
  // connect to node (0,0), don't use http!!!!
  port(DIR.IP('udp://localhost:10000'));
  // 0,0 node, udp port
  connect(DIR.IP('udp://localhost:10002'));
})
later(1500, () => {
  // Start agent here; it will migrate to cluster node (0,0)
```

```
  create('circulator',4);
})
```

*Ex. 14. Using the JAM shell to inject agents by extending code from Def. 15. A
simple circulator agent is created and started on the JAM shell control instance
(starting and controlling the cluster, but not being part of the cluster network).*

Virtual worlds can be constructed by using the JAM shell with
only a few operations, shown in Def. 16. Although, there is no
speed-up that can be created by virtual worlds (in contrast, to clus-
ters), virtual worlds allow testing of agent mobility in a container
world.

```
// 1. Create new Nodes (root node exists alreays
var nodes = [{x:1,y:0},{x:1,y:1},{x:0,y:1}].map(add);
// 2. Connect nodes with virtual circuit links
[{x1:0,y1:0,x2:1,y2:0},
 {x1:1,y1:0,x2:1,y2:1},
 {x1:1,y1:1,x2:0,y2:1},
 {x1:0,y1:1,x2:0,y2:0}].map((link) => {
 connect({x:link.x1,y:link.y1},{x:link.x2,y:link.y2})
});
```

*Def. 16. Creation of virtual linked JAN node worlds by the JAM shell. The JAM
creator instance is part of the virtual network (root node at logical position 0,0)*

Finally, a circulator agent is shown in Ex. 15, slightly changed
from the IP-based cluster example. The agent must prevent going
back in the opposite direction from which the agent has arrived.

```
function circulator(maxhop) {
  this.maxhop=maxhop;
  this.hops=0;
  this.last=null;
  this.goto=null;
  this.act = {
    start: () => { log('Start on '+myNode()); sleep(1000); },
    move : () => {
```

```
        var next,back=this.goto?DIR.opposite(this.goto):'';
        if (link(DIR.EAST) && DIR.EAST!=back) next=DIR.EAST;
        else if (link(DIR.SOUTH) && DIR.SOUTH!=back) next=DIR.SOUTH;
        else if (link(DIR.WEST) && DIR.WEST!=back) next=DIR.WEST;
        else if (link(DIR.NORTH) && DIR.NORTH!=back) next=DIR.NORTH;
        log(next)
        this.goto=next;
        if (next) moveto(next);
      },
      hello : () => { this.hops++; log('I am on '+myNode()) },
      end   : () => { log('END'); kill() },
  }
  this.trans = {
      start : move,
      move  : () => { return this.goto?hello:end },
      hello : () => { return this.hops<this.maxhop?move:end }
  }
  this.next='start';
}
compile(circulator,{verbose:false});
create('circulator',4);
```

*Ex. 15. A simple circulator agent is created and started on the root JAM shell instance*

### 5.2.5 AIOS Extension

The AgentJS AIOS, i.e., the AgentJS API for agents for the respective four privilege levels, can be easily extended. The host application can directly control agents via AIOS extension functions, shown in Def. 17. Any number of arguments that can be passed to extension functions is supported. Each agent with an AIOS containing the extension function can call the extension function in activities, transition rule functions, and signal handler functions. There are synchronous and computational functions, and asynchronous function with agent process control.

Simple computational functions can be added to the AgentJS AIOS for all or specific levels by using the `extend` function and by providing the AIOS level/levels, the new function name in the AIOS, and optionally the number of expected arguments (required only for the agent compiler to check code). New functions added to the AIOS can provide agent callback support. The callback functions will not be called in the agent object context (default case), i.e., the *this* object does not point to the agent object.

Additionally, the extension function may not start asynchronous function calls with callbacks (like `setInterval`) that could be called later outside of the agent activity in which the extension function was called. Additional logic is required. Direct agent process control is possible in the extension function via the global `Aios.current.process` variable. The process object supports the *suspend*, *wakeup*, and *kill* methods. Asynchronous extension functions can send a signal to the calling agent process, too. A signal can be added to the process object *signals* list attribute and by emitting a schedule event for this node (by the *node* attribute of the process).

In any out-of-flow processing of an asynchronous extension function the state of the agent must be checked by checking the *kill* and *dead* process object attributes first.

```
 1: extend : function (level:number|number[],
 2:                     name:string,
 3:                     function,
 4:                     argn?:number|number[])
 5:
 6: // 1. New pure computational and
 7: //    synchronous extension function
 8: function computation(data,callback) {
 9:    callback(ε(data)); // callback has no agent object context
10:    return ε(data)
11: }
12: // 1b. Calling a callback function in its agent context
13: function computation(data,callback) {
14:    var pro = Aios.current.process;
15:    // callback is executed in the current agent object context
16:    pro.callback(callback,ε(data));
17:    return ε(data)
18: }
```

```
19: // 2. Asynchronous extension function with
20: //     agent control (suspend/wakeup)
21: function control(callback) {
22:   // save calling agent process
23:   var pro = Aios.current.process;
24:   // Suspend agent
25:   pro.suspend(0);
26:   // Asynchronous function with
27:   // callback that is called outside
28:   // of this function
29:   asynchronous(() => {
30:     // Important: check agent state before proceed!
31:     if (pro.kill || pro.dead) return;
32:     // wake-up agent process
33:     pro.callback(callback,arg);
34:     pro.wakeup();
35:   });
36: }
37: // 3. Asynchronous extension function with
38: //     agent control (signal).
39: //     Agent must provide a signal handler, i.e.,
40: //     this.on = { 'SIGNAL': function () {..} }
41: function signal(sig,arg,..) {
42:   var pro = Aios.current.process;
43:   asynchronous(() => {
44:     // Check agent state first!
45:     if (pro.kill || pro.dead) return;
46:     pro.signals.push([sig,arg]);
47:     // Fire schedule event for this node
48:     Aios.emit('schedule',pro.node);
49:   });
50: }
51: // Add to AIOS level 1
52: extend(1,'computation',foo,2);
53: // Add to all AIOS levels
54: extend([0,1,2,3],'computation',foo,[1,2]);
55: // The computation function can now be used by all agents
56: extend([2,3],'control',control,1);
57: // The control function can now be used by
58: // all level-2 and level-3 agents
59: extend([1,2,3],'signal',signal,2);
```

*Def. 17. JAM shell AIOS extension API and programming templates*

### 5.3  JAM Web Laboratory

The JAM Web laboratory is a self-contained HTML-JS application for the Web browser using the browser version of the JAM shell providing basically the same API as the terminal command line version. The JAM laboratory provides shell, editor, monitor, and report windows (see Fig. 12). Single agents can be debugged or terminated by a tree-based agent explorer. Code text can be loaded and saved on the local file system via a tiny helper service *wex* executed on the local computer or alternatively via the Web browser file dialogue.

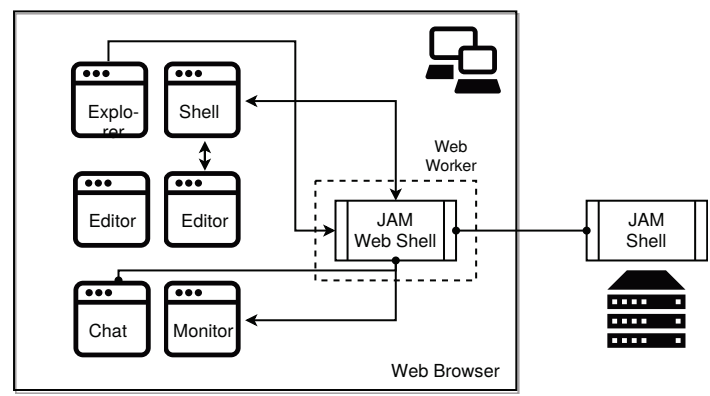

*Fig. 12. Software architecture of the Web browser JAM laboratory*

The JAM instance is processed in an isolated Web worker process not blocking the GUI. In addition, multiple physical JAM instances can be started and controlled via the GUI.

### 5.4  JAM Mobile App

There is mobile App version providing a full JAM instance with a dedicated GUI. The GUI is page-based  (see Fig. 13) and consists of configuration and network set-up, network management (connecting the JAM node), agent management, and life information pages displaying all available sensors, and a chat dialogue page. There is a pure HTML-CSS-JS and a Cordova framework version. The basic HTML version can be used in any Web browser on mobile and desktop computers. On mobile devices, the basic HTML version provides access to device sensors. The location of the device can be requested via GPS sensing (if available) and/or via the public IP address resolved by an external IP-GEO data base and by using the Mozilla location service (optional).

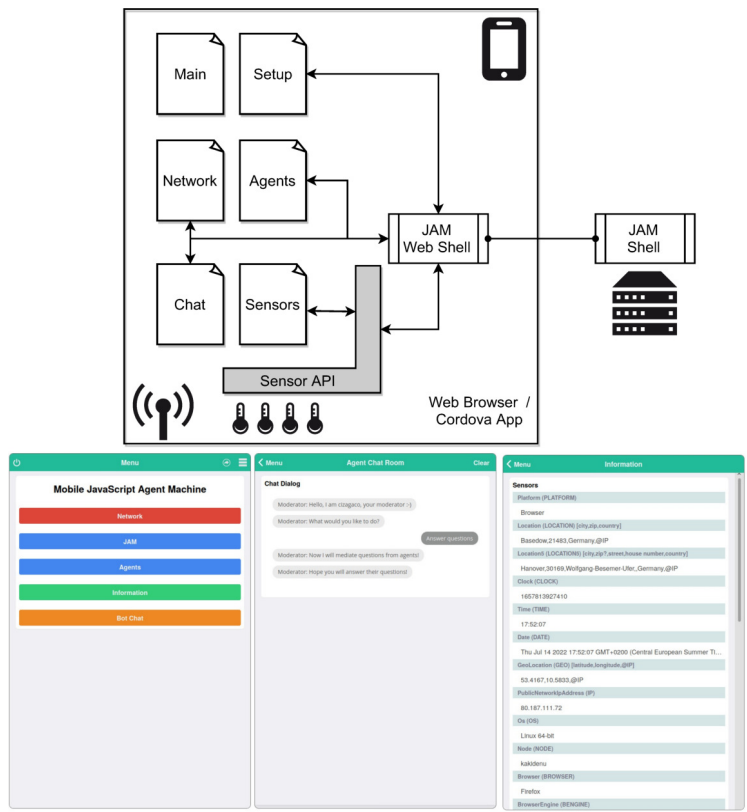

Fig. 13. *(Top) Software architecture of the Web browser / Cordova JAM App (Bottom) Some example pages*

A root agent can be loaded from a Web server or can be embedded already in the mobile App (a single self-contained HTML file or a multi-file archive). Additional agents can be loaded from a Web server and started. There is a JAM logging page collecting all agent and JAM messages. The chat dialogue can be accessed via an extended AIOS API by the agents directly (with sufficient rights). The agents can access the device sensors via the tuple space. The sensor information page lists all available sensors and their names used as the first key element of the sensor tuples (provided by a provider function).

The mobile App is used for mobile crowdsensing (MCS) applications as well as for education.

## 6. Performance

The JAM is typically processed by modern JS VMs like node.js/V8 or SpiderMonkey. These JS VMs can achieve native code performance of about 20-50%, mostly by applying just-in-time (JIT) native code compilation of already compiled Bytecode (the most common JS VM intermediate format). Although, JAM processing agents is written in JavaScript and processed by the underlying JS VM (pseudo VM in VM architecture), there is nearly no performance degradation in the processing of agents. The tasks with the highest computational costs are creation of agents and migration, both requiring full code-to-text serialisation and text-to-code back deserialisation. Fortunately, agent creation (not forking) requires this process only one time applied to a reusable agent constructor function.

### 6.1 JS VM

JAM performance is inherently bound to the underlying JS VM performance. In Tab. 15, results for different JS VMs and host platforms using a normalised dhrystone benchmark test are shown. The dhrystone benchmark utilises an average mix of operations typical for real-world software. i.e., object creation and destruction, array and string operations, and function calls. Different JS VMs are com-pared, although, JAM basically runs out of the box only on *node.js* and *jx*, a multi-threaded *node.js* derivative optimised for JAM (especially native agent process watchdog support) based on the *jx-core* project.

| JS VM | Host | dhrystones/s | Memory |
|-------|------|--------------|--------|
| C/native | x86 i5-4310U, 2GHz, 8GB | 23000k | - |
| quickjs | x86 i5-4310U, 2GHz, 8GB | 142k | 2MB |
| quickjs | Raspberry Pi Zero, Arm, 1GHz, 1GB | 9k | 1MB |
| node.js-8 | x86 i5-4310U, 2GHz, 8GB | 6800k | 28MB |
| node.js-8 | Raspberry Pi Zero, Arm, 1GHz, 1GB | 220k | 28MB |
| node.js-10 | x86 i5-4310U, 2GHz, 8GB | 8000k | 30MB |
| jx (node.js-0.10.4) | x86 i5-4310U, 2GHz, 8GB | 7000k | 18MB |
| SpiderMonkey (Gecko FF 78) | x86 i5-4310U, 2GHz, 8GB | 5330k | - |

*Tab. 15. Comparison of different JS VMs and host platforms using the normalised dhrystone benchmark test [32]*

Note that node.js and the V8 core engine were optimised for the x86 processor architecture and therefore underperform on the Arm processor architecture. Beside the computational power, the memory requirements (base memory required for the VM itself and JAM) is another important parameter in the deployment of (tiny) embedded systems like the Raspberry Pi Zero. The VM perfomance is about 4 times slower than the native (C) code performance of the beanchmark test program. Node.js with the V8 core engine with native code JIT compilation outperforms the low-resource pure Bytecode-based quickjs engine. But SpiderMonkey delivers similar results than V8!

## 6.2  Agent class compilation and agent creation

The experiment was performed with the *jamsh* program to evaluate the time required to compile agent class constructor functions and to create agent by them using the following code:

```
function foo(p) { .. }
N=1000;
t0=time()
for (var i = 0;i<N;i++)
  compile(foo,{verbose:false});
t1=time()
print((t1-t0)/N,'ms/compile')
N=100
t0=time()
for (var i = 0;i<N;i++)
  create('foo',{..});
t1=time()
print((t1-t0)/N,'ms/create')
var aid = create('foo',{..}),
    agobj = agent(aid);
print(Aios.Code.size(agobj))
```

The results are shown in Tab. 16. The compile time is basically neglectible because an agent class constructor must only be compiled one time (per node). The creation of agents is fast enough (at least on modern computer hardware) compared with a typical average agent creation rate (about 1/s-10/s).

| Operation | Parameter | Time / Aux. |
|-----------|-----------|-------------|
| compile | function | 0.8 ms |
| create | { string[16], number } | 0.09 ms |
| size | agent | Ser. code: 365 Bytes |

*Tab. 16. JAM (jamsh) performance: Agent sandbox-restricted transformation and creation (Host: x86 i5-4310U, 2GHz, 8GB, jx/node.js 0.10.40)*

## 6.3 Migration

Firstly, a $2 \times 2$ virtual JAM node mesh-grid network was created. The experiment was performed with the *jamsh* and the following test code:

```
function foo(p) {
  ..
  this.act = {
    start : ..
    percept : () => {
      this.goto=null;
      var dirs = [DIR.NORTH,DIR.SOUTH,..]
      iter(dirs, () => {
        if (!this.goto && link(dir))
          this.goto=dir;
      })
    },
    move : () => {
      this.hops--;
      moveto(this.goto)
    },
    end : ..
  }
  this.trans = {
    start:percept,
    percept : () => this.goto && this.hops?move:end,
    move : percept
  }
}
```

```
var node2 = add({x:1,y:0}),
    node3 = add({x:1,y:1}),
    node4 = add({x:0,y:1});
// unidirectional links!
connect({x:0,y:0},{x:1,y:0})
connect({x:1,y:0},{x:1,y:1})
connect({x:1,y:1},{x:1,y:0})
connect({x:1,y:0},{x:0,y:0})

compile(foo,{verbose:false});
var N=1000
create('foo',{data:'..',hops:N});
start()
```

The second experiment was performed with a physical 2 × 2 node mesh-grid cluster using directional UDP links and the following code:

```
function foo(p) {
  ..
  this.act = {
    start : () => {
      // goto cluster node(0,0)
      var connected=link(DIR.IP('%'))
      if (connected && connected.length)
        this.goto=DIR.NODE(connected[0]);
    },
    percept : () => {
      this.goto=null;
      // follow ring
      var dirs = [DIR.EAST,DIR.SOUTH,DIR.WEST,DIR.NORTH];
      var next = dirs[this.index];
      if (link(next)) {
        this.goto=next;
        this.index=(this.index+1)%4;
      }
    },
    move : () => {
      this.hops--;
      moveto(this.goto)
    },
    end : ..
  }
  this.trans = {
    start   : () => this.goto?'move':'percept',
    percept : () => this.goto && this.hops?'move':'end',
    move    : 'percept'
```

```
  }
}
var workers = cluster({
  connect:true,
  rows:2,
  cols:2,
  port0:11001,
  port1:10001,
  portn:100,
  proto:['http','udp'],
  poll:2000,
  todo:'start()',
  verbose:0,
});
workers.start()
// Connect to cluster node (0,0)
var p1=port(DIR.IP('udp://localhost:12000'));
later(500,() => {
  connect(DIR.IP('udp://localhost:10002'))
});
compile(foo,{verbose:false});
var N=1000
later(3000,() => {
  create('foo',{data:..,hops:N});
})
```

The following results in Tab. 17 contain average times for agent process migration. The migration includes agent data and code serialisation, transfer of the serialised code to the next node, and finally the deserialisation and process creation on the new node. The migration between virtual nodes is as fast as agent creation, whereas agent migration between physical nodes is 10 times slower, i.e., the overhead of AMP and the communication network is significant.

| Operation | Parameter | Time / Aux. |
|-----------|-----------|-------------|
| moveto | Virtual Link[1] | 0.5 ms / hop |
| moveto | Physical UDP Link[2] | 6.0 ms / hop |
| size | agent[1] | Ser. code: 730 Bytes |

| size | agent$^2$ | Ser. code: 946 Bytes |
|------|-----------|----------------------|

*Tab. 17. JAM (jamsh) performance: Agent migration (Host: x86 i5-4310U, 2GHz, 8GB, jx/node.js 0.10.40)*

## 6.4 Communication

Communication is performed between a parent and child agent created by forking. The child agent migrates to another node (virtual or physical). Note that the agents in communication experiments use a ping-pong flow with signal handlers and activity blocking, i.e., agent scheduling is included in the communication times.

The first signal experiment used the following code skeleton using two agents on two different nodes (virtual or physical):

```
function foo(p) {
  ..
  this.act = {
    start : () => {
      this.child=fork({parent:me(),goto:DIR.EAST});
    },
    move : () => {
      moveto(this.goto);
    },
    ping : () => {
      this.count--;
      if (this.parent) send(this.parent,'PING',time());
      else send(this.child,'PING',time());
    },
    pong : () => { sleep() },
    wait : () => { sleep() },
    end  : () => {  kill() },
  }
  this.trans = {
    start : () => { return this.goto?'move':'wait' },
    wait : () => { this.count--;
                   this.time=time(); return ping },
    move : ping,
    ping : pong,
    pong : () => { return this.count?'ping':'end' },
```

```
  }
  this.on = {
    PING : (arg,from) => {
      send(from,'PONG',time())
      if (this.child && this.count==this.N) wakeup()
    },
    PONG : (arg,from) => {
      wakeup()
    },
  }
  ...
```

The second agent is forked from the first one. The tuple space communication can only be executed between agents on the same JAM node. Again, two agents are used in the test set-up. Results are shown in Tab. 18. Signal sending is as fast as tuple space communication. Remote signal propagation is affected by the communication overhead, too, approximately 30 times slower.

| Operation | Parameter | Time / Aux. |
|-----------|-----------|-------------|
| send | signal, virtual nodes | 0.1 ms |
| send | signal, physical nodes | 3.5 ms |
| out,inp | one node | 0.03 ms |

*Tab. 18. JAM (jamsh) performance: Agent communication times using signals and tuples including agent scheduling times (Host: x86 i5-4310U, 2GHz, 8GB, jx/node.js 0.10.40)*

## 7. SEJAM

### 7.1 Concept and Architecture

The Simulation Environment for JAM (SEJAM) is a graphical extension on the top of the JAM platform to support agent-based simulation (ABS) using the same AgentJS programming as used in agent-based computation (ABC). Moreover, ABS and ABC can be com-bined connecting SEJAM with external JAM nodes. Agents in simulation can migrate between "virtual" and real worlds seamlessly. The basic SEJAM architecture and concept is show in Fig. 14. A typical example of combining ABS with ABC is simulation with data from real environments using mobile crowd sensing, e.g., as described in [14].

Graphical shapes in the 2D simulation world are coupled to virtual JAM instances as part of the virtual node world of JAM. A visual shape is typically processing one agent, e.g., an agent representing a digital twin of a human or a machine. But an unlimited number of agents can be processed by one virtual node.

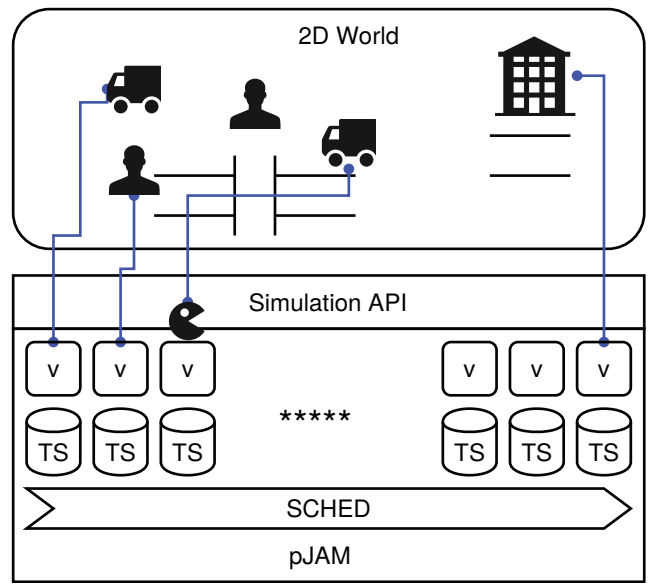

*Fig. 14. SEJAM concept and architecture: A graphical simulation world on top of a JAM node instance*

There are basically to different sub-classes of agents in the simulation:

1. Physical agents representing physical entities (humans, machines, animals, resources) as an ABS cell;

2. Computational agents that are used for distributed computation in the simulation and real world as an ABC cell.

Both agent sub-classes are processed by the same JAM instance. But physical agents are bound to their virtual node and cannot migrate. Additionally, they can access a NetLogo similar simulation programming interface, e.g., supporting the *ask* operator to explore agents in the spatial neighbourhood in the simulation world [30].

A physical agent consists of v-JAM node and one body agent always bound to this node. Physical agents can perform any computation and communication with other v-JAM nodes. They can create computational agents that are capable to migrate to other v-JAM nodes of physical agents or external nodes. For example, a physical node can define a multicast port (used in simulation, only) with a visual shape of a circle with a specific radius. If there are two physical nodes with overlapping communication circles, the AIOS `link(DIR.PATH('%'))` function will return the node name of the neighbour node. The physical agent can create a computational agent that is then capable to migrate to the neighbouring node.

## 7.2 Simulation Model and API

### 7.2.1 Simulation World

The simulation model is typically one JS (JSON-like) file containing the model object and the agent class constructor functions with its visual descriptors. Alternatively and for the sake of readability, the agent class constructor functions can be loaded from separate files containing the (named) agent class constructor function, shown in Def. 18. Examples of agent classes are shown in Appendix A. Beside generic two-dimensional graphical worlds, different pre-defined world architectures can be used:

1. Mesh-grid consisting of a two-or three-dimensional connected grid of v-JAM nodes with neighbourhood connectivity (computational agents only), basically used to study distributed data processing and communication systems (see Def. 19). Network v-JAM odes are connected with von-Neumann neighbourhood nodes providing DIR.NORTH, DIR.SOUTH, DIR.WEST, and DIR.EAST directional communication links.

2. Patch-grid world consisting of a two-dimensional Cartesian map of patches supporting physical agents (i.e., container of a coupled v-JAM with one agent and a visual shape). A patch can be populated with one or more agents. Agents can move freely to different patch positions together with their container. A NetLogo-like API is provided that can be used by physical and world agents (see Def. 18).

```
 1: {
 2:   name:string,
 3:   // Definition of agent classes
 4:   agents : {
 5:     classname: {
 6:       behaviour : (..) => { .. } |
 7:                   open(filename.js),
 8:       visual : { shape, width, height, fill, .. }
 9:       ..
10:   },
11:   // some passive resources in the 2D simulation world
12:   resources : {
13:     classname : (id,x,y,w,h) => {
14:       return { id, class, visual }
15:     },
16:     ..
17:   },
18:   // dynamic node constructors, mainly for physical
19:   // agent nodes and the world node
20:   nodes : {
21:     classname : (x,y) => {
22:       return { id, x, y, visual, ports }
23:     },
```

```
24:      ..
25:  },
26:  // Definition of the world
27:  world : {
28:    init : {
29:       // create specific agents
30:       agents : {
31:         classname : (nodeid) => {
32:           return { level, args }
33:         },
34:       }
35:    },
36:    // create special nodes
37:    map : () => {
38:      return [
39:        model.nodes.world(52,52),  // graphical position
40:      ]
41:    },
42:    // resources
43:    resources : (model) => {
44:       // using model.resources.classname functions
45:      return {} []
46:    },
47:    // if there is a NetLogo patchgrid world
48:    patchgrid : {
49:      rows, cols, width, height, floating?
50:    },
51:  },
52:  // simulation parameter
53:  parameter? : { .. },
54: }
```

*Def. 18. Basic template of a SEJAM2 simulation model supporting the patch-grid architecture*

The patch-grid world is used in conjunction with the NetLogo API that can be used by physical agents and the world agent. A patch-grid is a discretised 2D world with equally sizes patches. Most operations use the logical path coordinates instead of the graphical pixel coordinates (used otherwise). Instead of logical or pixel coordinates, the world and the visual shapes can be referenced by geospatial coordinates using `gps:{latitude, longitude,`

`height`} values (e.g., in the visual node descriptors). A geographical coordinates mapping function is required.

The simulation commonly uses one central world agent (agent class "world") to control the simulation globally and centrally. The world agent has full access to JAM and SEJAM APIs including data base and file system access.

The agent simulation world can be coupled a multi-body physics simulator (CANNON, see [15] for details).

```
1:  {
2:    ..
3:    world : {
4:      init: {
5:        agents: {
6:          node:function (nodeId,position) {
7:            // Create on each node a node agent,
8:            // return respective agent parameters!
9:            // If no agent should be created on the
10:           // respective node, undefined must be returned!
11:           if (nodeId!='world')
12:             return {level:2,args:[
13:               {x:position.x,y:position.y,z:position.z,
14:                model:model1,
15:                verbose:0,
16:               }]}
17:         },
18:         world: function(nodeId) {
19:           if (nodeId=='world') return
20:             {level:3,args:[{model:model}]};
21:         }
22:       }
23:     },
24:     meshgrid : {
25:       // y-axis
26:       rows:8,
27:       // x-axis
28:       cols:5,
29:       //z-axis
30:       levels:3,
```

```
31:
32:       matrix:[[0,0],[250,0], [500,0]],
33:
34:       node: {
35:         // Node ressource visual object
36:         visual : {
37:           shape:'rect',
38:           width:30,
39:           height:30,
40:           fill: {
41:             color:'green',
42:             opacity: 0.5
43:           }
44:         },
45:         filter: function (pos) {
46:           // Node filter to create irregular networks
47:           return
48:         },
49:       },
50:       // Link port connectors
51:       port: {
52:         type:'unicast',
53:         place: function (node) { return [
54:           {x:-15,y:0,id:'WEST',',
55:           {x:15,y:0,id:'EAST'},
56:           {x:0,y:-15,id:'NORTH'},
57:           {x:0,y:15,id:'SOUTH'},
58:           {x:-15,y:-15,id:'UP'},
59:           {x:15,y:15,id:'DOWN'}
60:         ]},
61:         visual: {
62:           shape:'rect',
63:           fill: {
64:             color:'black',
65:             opacity: 0.5
66:           },
67:           width: 5,
68:           height: 5
69:         }
70:       },
71:       // Connections between nodes
72:       // (with virtual port connectors)
73:       link : {
74:         // Link resource visual object
75:         connect: function (node1,node2,port1,port2) {
```

```
76:          return true},
77:        type:'unicast',    // unicast multicast
78:        visual: {
79:          shape:'rect',
80:          fill: {
81:            color:'#888',
82:            opacity: 0.5
83:          },
84:          width: 2
85:        }
86:      }
87:    }
88:  }
89: }
```

*Def. 19. Mesh-grid world architecture defining a communication network with von-Neumann neighbourhood connectivity consisting of v-JAM nodes processing computational agents*

The difference between the mesh-grid communication network supporting computational agents only and the patch-grid world supporting physical (and computational) agents is shown in Fig. 15.

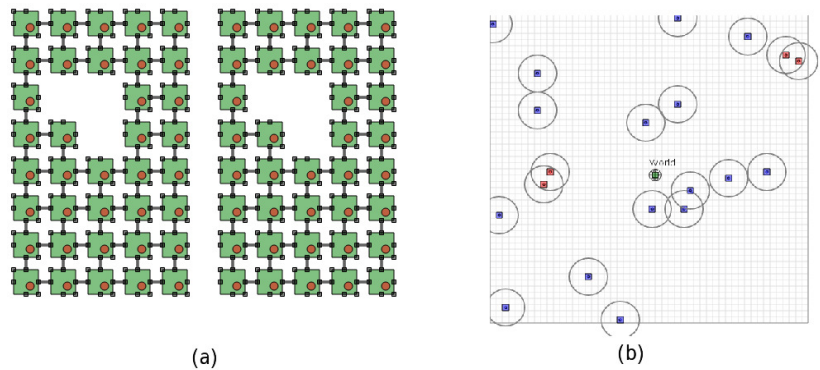

(a)                           (b)

*Fig. 15. Difference between the mesh-grid (a) and patch-grid world (b)*

### 7.2.2 *NetLogo API*

SEJAM provides a NetLogo-like API for physical agents. NetLogo is a widely used agent-based simulation environment for studying complex dynamic systems. Details of NetLogo simulations can be found in [31]. The API is summarised in Tab. 19 and can be accessed by the *net* module extension. If arrow functions are supplied the `this` agent self-reference is not available. Instead, an additional callback function argument provides the agent object self-reference. There are different simulation object classes: Agents, resources, and patches, which can be handled by the `net.ask` function.

| Operation | Parameter | Description |
|---|---|---|
| `net.create` | agent-*agentclass*, *number*, *callback*(*self*, *index*)? | Creates a set of agents. Each created agent can be configured by the callback function using the *self* function parameter pointing to the agent object. |
| `net.setxy` | [*x*, *y*] | Set patch position of agent. |
| `net.ask` | agent or agent-*agentclass*, [*x*, *y*], *callback*(*self*)? | Returns all agents at grid position (*x*, *y*) and optionally iterates over all agent objects with a callback function. |
| `net.ask` | agent or agent-*agentclass*, *radius*:number or `*`, *callback*(*self*)? | Returns all agents around current agent position in given radius or all. |

| net.ask | resource or resource-*class*, .. | Returns all matching resources |
|---------|----------|------------|
| net.ask | patch, .. | Returns all matching patches |
| net.ask | *objclass* or *objclass-class*, {x,y,w,h} or {x0,y0,x1,y1}, *callback(self)*? | Returns all objects of given object class in given bounding box. |
| net.set | color or shape, *value* | Set visual shape colour or shape (circle, rect, triangle) |

*Tab. 19. SEJAM NetLogo-like API for physical agents*

## 7.3 Software

There are two nearly identical versions of the *sejam2* simulation environment:

1. A *node-webkit* version with full access to operating system services and networking;

2. A Web browser version with limited access to operating system services.

The original *sejam* simulator was a terminal and command line implementation using *ncurses* for a textual 2D world visualisation.

## 7.4 Example

A complete example of a simple simulation model is shown in Appendix A. The simulation model consists of:

1. The patch-grid world providing the NetLogo API;

2. Computational world agent;

3. Physical agents performing random walk in the simulation world and changing their node color based on crowd perception.

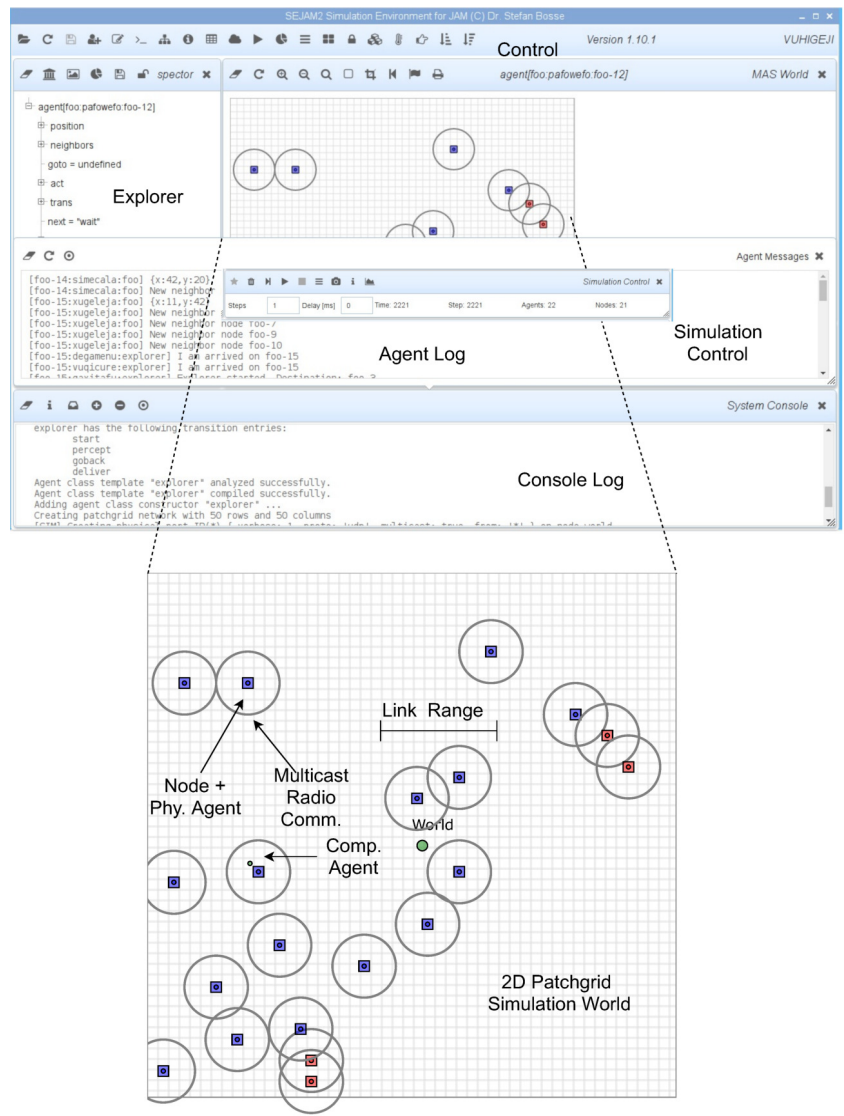

*Fig. 16. SEJAM2 GUI window (node-webkit version) with example world*

The following use-case section will elaborate a more complex example.

## 8. Use case: Mobile Crowd Sensing

An extended use case deploying mobile agents in an ad-hoc mobile crowd sensing (MCS) framework should demonstrate the suitability and scalability of the JAM architecture and its software. Agent-based methods with an unified agent model features:

- Mobile crowdsensing that is used to sample environmental and user data on micro scale level;

- Tight coupling of simulation (ABS) with real world (human-in-the-loop);

- Incremental data collection by software agents synchronises simulation with real world;

- Simulation snapshots and forking enables prediction of future world evolution.

### 8.1 The Concept

The basic concept relies on long-term longitudinal data mining combining MCS, modelling, and simulation [20]. The principle system architecture is shown in Fig. 17. In addition, details can be found in [10] and [14] where MCS is integrated in the agent-based simulation using SEJAM.

The mobile crowd sensing framework deploys the following nodes:

1. Stationary JAM relay nodes with a public internet port and access to data base services;

2. Web browser JAM shells (JAM lab) with GUI for data access and analysis;

3.  Web browser JAM nodes embedded in Web pages.

4.  Mobile Apps with JAM shell (Web browser and Android App versions)

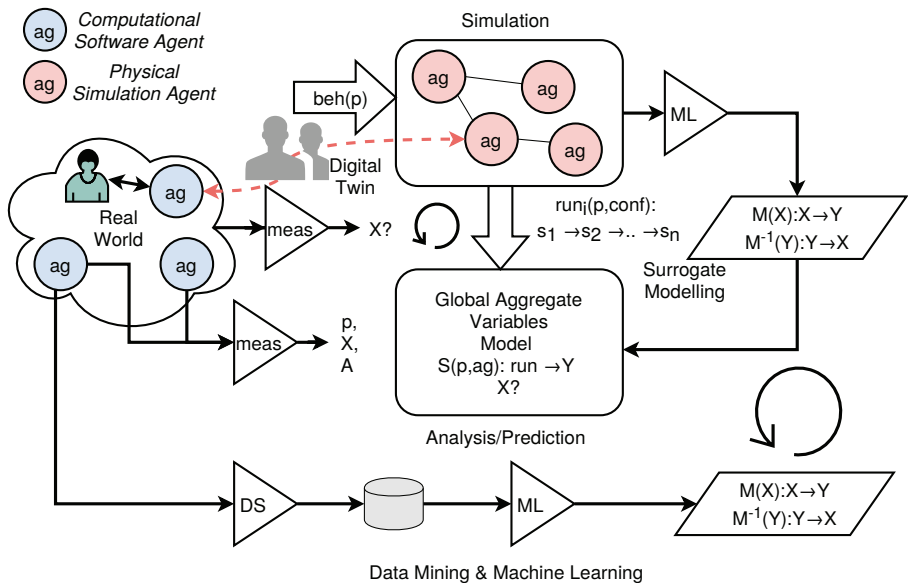

*Fig. 17. Unified agent methodology for longitudinal data mining, modelling of complex systems, and simulation*

## 8.2 The MAS

The augmented simulation MAS consists of:

1.  A world agent controlling the simulation and managing data in SQL data bases;

2.  Computational survey explorer agents that are created by the world agent;

3.  Static set of physical agents modelling simple social and networking behaviour based on a-priori model parameter sets;

4. Dynamic set of physical digital twin agents with model parameter settings based on micro surveys and created by the world agent via the explorer agent feedback.

The survey agents are started in the simulation world and migrate to a public relay JAM node. Mobile devices (smartphones) using the mobile JAM App are connect to the same relay node (ad-hoc). The explorer agent stays on the relay for a specific time and watches for new connections. If there is a new connection, a child agent is forked and sent to the remote node. The child explorer agent tries to find a chat agent. If the chat agent was found, the micro survey is delegated to the chat agent. The explorer agent waits for a specific time for response (by the user propagated by the chat agent). Communication with the chat agent is performed via the tuple space.

A simplified version of the MAS is shown in Appendix A.

## 9. Conclusion

The JavaScript Agent Machine (JAM) is an universal and highly portable agent processing platform using an advanced VM architecture. JAM is entirely programmed in JavaScript (JS) and can be executed on any JS VM engine, e.g., *node.js* or *SpiderMonkey*. The agents are programmed in AgentJS, a sub-set of JS. The agent behaviour is composed of activities and transitions between activities based on body variables of the agent. JAM agents are mobile processes that can migrate a process snapshot including their code between different JAM nodes. JAM supports virtualisation on different levels including agent communication. JAM can be deployed in Web-based and Internet environments like the IoT. JAM addresses ABS as well as ABC, combined in the SEJAM simulator. Major fields of application are general distributed computing in the Internet and Web, simulation, sensor processing in distributed sensor networks, and mobile crowd sensing. The Agent Management Port (AMP) protocol enables loosely coupled connection between JAM nodes using a wide range of communication protocols.

## 10. Appendix A.

A simplified version of the MCS-MAS is shown in this appendix combining ABS with ABC performing MCS to create digital twins in the simulation world from survey results.

### 10.1  Example of a SEJAM2 Simulation Model

In Ex. 16, a simple SEJAM2 simulation model is shown. The model is composed of three different agent models (a world agent controlling the simulation, a physical simulation agent, and a computational agent), defined in the *agents* section in lines 4-44. The agent code is loaded from separate files. Each agent class is associated with the code behaviour and a visual shape. Agents require v-JAM nodes for execution typically associated with a visual shape and a position in the simulation world, defined in the *nodes* section in lines 47-105. Finally, the simulation world (here the pre-defined patch-grid) is defined in lines 113-139, including initialisation and mapping functions. The *map* function returns a list of node descriptor structures. The *init* object is primarily used to create an initial static set of agents. Each entry in the *init.agents* field specifies an agent class name that must be defined in the *agents* section. Typically, agents are created dynamically by the world agent at simulation time.

The world node creates a physical AMP communication endpoint (only possible in the node-webkit SEJAM version), defined in lines 48-51 and 71. It is used to connect the simulation world (the automatically created world node more specifically) with external JAM nodes. The link enables agent to migrate between simulation and real world nodes. Examples of a computational agent can be found in the Sec. 10.5

The *foo* class agents create virtual wireless radio communication ports which can be used to link neighbouring nodes (*port* entry in 89-102 lines ). Only overlapping radio circles enable the linking of nodes (automatically done by the simulator program), and the communication radius is given by the visual shape radius.

```
 1: {
 2:   name:'Random Walk Simulation',
 3:   agents : {
 4:     world : {
 5:       behaviour:open('world.js'),
 6:       visual:{
 7:           shape:'circle',
 8:           width:10,
 9:           height:10,
10:           fill: {
11:             color:'green',
12:             opacity: 0.0
13:           }
14:       }
15:     },
16:     foo : {
17:       behaviour : open('agent.js'),
18:       visual:{
19:           shape:'circle',
20:           width:4,
21:           height:4,
22:           fill: {
23:             color:'blue',
24:             opacity: 0.0
25:           }
26:       },
27:       type:'physical',
28:     },
29:     explorer : {
30:       behaviour : open('explorer.js'),
31:       visual:{
32:           shape:'circle',
33:           width:4,
34:           height:4,
35:           x:-8,y:-8,
36:           fill: {
37:             color:'green',
38:             opacity: 0.0
39:           }
40:       }
41:     },
42:   },
43:   resources : {
44:
45:   },
46:   nodes: {
47:     world: function (x,y) {
```

```
48:          var phy={type:'physical',
49:                   ip:'*',
50:                   to:model.parameter.ip+':'+model.parameter.ipport,
51:                   proto:model.parameter.proto};
52:          return {
53:            id:'world',
54:            x:x, // patch position
55:            y:y,
56:            visual : {
57:              shape:'icon',
58:              icon:'world',
59:              label:{
60:                text:'World',
61:                fontSize:14
62:              },
63:              width:20,
64:              height:20,
65:              fill: {
66:                color:'black',
67:                opacity: 0.5
68:              }
69:            },
70:            ports : {
71:              phy:phy
72:            }
73:          }
74:      },
75:      foo : function (x,y,id) {
76:          return {
77:            id:id,
78:            x:x, // patch position
79:            y:y,
80:            visual : {
81:              shape:'rect',
82:              width:10,
83:              height:10,
84:              fill: {
85:                color:'yellow',
86:                opacity: 0.5
87:              }
88:            },
89:            port: {
90:              type:'multicast',
91:              status: function (nodes) {
92:                // Filter out nodes, e.g., beacons only?
93:                return nodes;
94:              },
95:              visual: {
96:                shape:'circle',
97:                radius : 30,
```

```
 98:              line: {
 99:                color: 'grey'
100:              }
101:            }
102:          },
103:
104:        }
105:      }
106:  },
107:  parameter : {
108:    ip:'ag-0.de',
109:    ipport:10002,
110:    proto:'udp',
111:    phy:true,
112:    speed: 0.2,
113:    randomness: 0.1,
114:  },
115:  world : {
116:    init: {
117:      agents: {
118:        world: function(nodeId) {
119:          if (nodeId=='world')
120:            return {level:3,args:{verbose:1}};
121:        }
122:      }
123:    },
124:    // special nodes
125:    map : function () {
126:      return [
127:          model.nodes.world(25,25), // patchgrid position
128:      ]
129:    },
130:    // resoirces
131:    resources : function () {
132:      return []
133:    },
134:    patchgrid : {
135:      rows : 50,
136:      cols : 50,
137:      width : 10, // geometrical width and height of patch in pixels
138:      height : 10,
139:      floating : true,  // physical agents are containers
140:                        // <logical v-node, behavioural agent>
141:    }
142:  }
143: }
```

*Ex. 16. Example SEJAM2 simulation model*

## 10.2 Example of a simulation world agent

The world agent of a simulation (always level 3) is responsible to control the simulation. It creates an initial set of agents, monitors simulation aggregate variables, and loads and saves data. The world agent creates an explorer survey agent that migrates to an external relay JAM node. Mobile devices (Smartphones) using the JAM App connect to the relay, too. The survey agent collects some user data and returns the result back to the world agent, which creates digital twin agents (random walk agent with a parameter set derived from the survey). The current simulation model (as defined in the previous section) can be accessed via the generic *simu* module and by calling the *model* function (line 7). The activity *percept* firstly checks for connected external JAM nodes. Secondly, it asks for all agents (nodes) of class *foo* and calculates an average (centre of mass) position from all nodes. Thirdly, it tests the presence of a survey reply tuple delivered by the survey agent. If a survey answer was found, a new digital twin agent is created with agent parameters derived from the survey results. Finally, a new instance of the survey explorer agent is created.

```
1:  function world(p) {
2:    this.modelSim=null;
3:    this.survey=null;
4:    this.act = {
5:     start : () => {
6:      log('World is started.');
7:      this.modelSim = simu.model();
8:      sleep(500)
9:     },
10:    wait : () => {
11:       sleep(500)
12:    },
13:    percept : () => {
14:     var nodes = link(DIR.IP('%'));
15:     log('Connected? '+nodes);
16:     var avgpos,n=0;
17:     net.ask('agents-foo','*', (agent,node) => {
18:      if (!avgpos) avgpos=copy(agent.pos);
```

```
19:     else {
20:       avgpos.x+=agent.pos.x;
21:       avgpos.y+=agent.pos.y;
22:     }
23:     agent.speed=this.modelSim.parameter.speed;
24:     agent.randomness=this.modelSim.parameter.randomness;
25:     n++
26:   })
27:   avgpos.x/=n;
28:   avgpos.y/=n;
29:   log('Center of mass: ',avgpos,n);
30:   // return from survey agent?
31:   if (test(['SURVEY','What kind of person are you?',_,_])) {
32:     var t = inp(['SURVEY','What kind of person are you?',_,_]);
33:     if (t[2]) {
34:       var answer = t[2][4],
35:           from  = t[2][1];
36:       var s=0.1,r=0.1;
37:       switch (answer) {
38:         case 'quiet': s=0.05; r=0.05; break;
39:         case 'restless': s=0.1; r=0.2; break;
40:         case 'prudent': s=0.2; r=0.05; break;
41:         case 'risky': s=0.5; r=0.2; break;
42:       }
43:       // create digital twin based on parameter set
44:       net.create('agents-foo',1, (agent,index) => {
45:         while (true) {
46:           var x=random(0,49),y=random(0,49);
47:           if (net.ask('agents-foo',[x,y]).length==0) {
48:             net.setxy(x,y)
49:             break
50:           }
51:         }
52:         agent.speed=s;
53:         agent.randomness=r;
54:         net.set('shape','triangle');
55:       })
56:     }
57:     // start new survey agent
58:     this.survey=create('survey',{},1);
59:   }
60: },
61: init : () => {
62:   // NetLogo API
63:   net.create('agents-foo',20, (agent) => {
64:     while (true) {
65:       var x=random(0,49),y=random(0,49);
66:       if (net.ask('agents-foo',[x,y]).length==0) {
67:         net.setxy(x,y)
68:         break
69:       }
```

```
69:      }
70:    })
71:    // create survey agent
72:    this.survey=create('survey',{},1);
73:   }
74:  }
75:  this.trans = {
76:   start:init,
77:   init:wait,
78:   wait:percept,
79:   percept:wait,
80:  }
81:  this.next = 'start';
82: }
```

*Ex. 17. Example SEJAM2 world agent (world.js)*

### 10.3 Example of a physical random walker agent

The physical simulation agent performs random walk in the simulation world. The speed and randomness parameter determine the walking behaviour. If an agent finds another agent in the neighbourhood it will change its colour to red, otherwise to blue. This is done in the activity *percept*. The *move* activity sets a new position of the agent if there is no other agent on the destination patch and if the position is inside the world bounding box. The walking direction is changed from time to time randomly in activity *percept* (lines 26-32).

In addition to the random walking behaviour, computational explorer agents are created that can migrate to neighbouring nodes (lines 35-46).

```
1: function foo(p) {
2:  this.position = null;
3:  this.neighbors = [];
4:  this.goto=null;
5:  this.speed=p.speed||0.1;
6:  this.randomness=p.randomness||0.1;
7:  this.act = {
8:   start : () => {
9:    this.position = simu.position();
10:    log('Agent is started  @ '+this.position.x+','+this.position.y)
```

```
11:  },
12:  wait : () => {
13:    sleep(200) // simulation steps
14:  },
15:  percept: () => {
16:    // NetLogo API
17:    var self=this,crowd=0;
18:    this.position = simu.position();
19:    net.ask('agents-foo',2, (agent,node) => {
20:      if (agent.agent==self.id) return;
21:      var dist = Math.max(1,distance(node.position,self.position));
22:      crowd += (1/dist);
23:    })
24:    if (crowd>0.1) net.set('color','red');
25:    else net.set('color','blue');
26:    if (random(0,1,0.1)<this.randomness) {
27:      var dx = random([-1,0,1]),
28:          dy = random([-1,0,1]);
29:      this.goto={
30:        x:dx,
31:        y:dy
32:      };
33:    }
34:    // Virtual wireless communication: simulated ad-hoc v-JAM connectivity..
35:    var neighbors=link(DIR.PATH('*'));
36:    // Some new nodes?
37:    iter(neighbors, (node) => {
38:      if (!contains(this.neighbors,node)) {
39:        log('New neighbor node '+node);
40:        this.neighbors.push(node);
41:        create('explorer',{
42:          goto:node,
43:          root:myNode(),
44:        },1); /* only level 1-2 agents are mobile! */
45:      }
46:    })
47:  },
48:  move: () => {
49:    // NetLogo API
50:    if (random(0,1,0.1)>this.speed) return;
51:    var dx = this.goto.x,
52:        dy = this.goto.y;
53:    if (this.position.x+dx<0 ||  this.position.x+dx>49) dx=0;
54:    if (this.position.y+dy<0 ||  this.position.y+dy>49) dy=0;
55:    // check new place occupancy
56:    var x = this.position.x+dx,
57:        y = this.position.y+dy;
58:    if (net.ask('agents-foo',[x,y]).length==0)
59:      net.setxy(x,y);
60:  },
```

```
61:  }
62:  this.trans = {
63:   start:wait,
64:   wait:percept,
65:   percept:() => { return this.goto?move:wait },
66:   move: wait,
67:  }
68:  this.next = 'start';
69: }
```

*Ex. 18. Example of simple physical simulation agent with random walk behaviour (agent.js)*

## 10.4 Example of a computational explorer agent

The following example of a simple computational explorer agent should only demonstrate the migration of this agent between different v-JAM nodes of the simulation world (using virtual *PATH* links only available in simulation). The destination node (its identifier name) is passed via the constructor parameter. The links between nodes are created dynamically by the simulator program (see simulation model). The explorer agent is created by the previously introduced physical random walker agent.

The explorer agent migrates to the destination node (activity *start*), senses the current clock in the destination node and its position in activity *percept*, going back to its root node in activity *goback*, finally delivering the sensor results in activity *deliver*.

```
1: function explorer(p) {
2:  this.goto=p.goto;
3:  this.root=p.root;
4:  this.sensor=null;
5:  this.act = {
6:   start: () => {
7:    log('Explorer started. Destination: '+this.goto);
8:    if (link(DIR.PATH(this.goto))) {
9:     log('Going to '+this.goto);
10:     moveto(DIR.PATH(this.goto));
11:    } else { log('No path to '+this.goto); this.goto=null };
12:   },
13:   percept : () => {
```

```
14:     log('I am arrived on '+myNode());
15:     this.sensor={time:clock(),pos:info('node').position};
16:     this.goto=null;
17:   },
18:   goback : () => {
19:     if (link(DIR.PATH(this.root))) {
20:       this.goto=this.root;
21:       log('Going to '+this.goto);
22:       moveto(DIR.PATH(this.goto));
23:     } else sleep(50);
24:   },
25:   deliver : () => {
26:     log('I am back on '+myNode()+' with sensor '+this.sensor);
27:   },
28:   end : () => {
29:     log('Terminating.');
30:     kill()
31:   }
32: }
33: this.trans = {
34:   start:() => { return this.goto?percept:end },
35:   percept : goback,
36:   goback  : () => { return this.goto?deliver:goback },
37:   deliver : end,
38: }
39: this.next='start'
40: }
```

*Ex. 19. Example of simple computational explorer agent (explorer.js)*

## 10.5 Example of a computational survey agent

The following more complex survey explorer agent features path travelling over external relay JAM nodes (public servers). The explorer agent tries to find mobile nodes with a chat agent (detected by the presence of a *CHAT* tuple). If the explorer agent stays on a mobile node with a chat agent (assuming a human user is available for questions), the agent will submit messages and questions to the chat agent via the local tuple space. The survey waits for a limited time on the answer, provided via the tuple space, too. In this version, the survey agent only visits one mobile node. It can be easily extended to visit multiple mobile nodes all connected to the relay node. The code of the chat moderator agent is shown in Sec. 10.6.

The chat agent (with direct access to the mobile App chat dialogue) processes client requests received via the tuple space. There are basically three different requests: Messages without an answer, question with selection from choices `{choices: [A,B,C,..], mutual?}`, and numerical and text input with a question `{type: 'text' | 'number', default?, value?: number | string}`.

The activity trace of the agent starts with the *search* activity. The *search* activity searches for externally connected nodes. If the *relay* body variable is not set, it will select the first found node and migrates to this node (assuming it is a relay node). It the agent is already on the relay node, it selects another node (not root origin node) to migrate to it. If no more node is found, it initiates the *goback* activity.

If the agent has found a node with a chat agent by testing the `CHAT` tuple in the *percept* activity, it will open a chat session and requesting a time-limited session token in the *sessionOpen* activity. If this was successful, it will ask his question in the activity *ask* and waits for the reply of the chat agent or a time-out. After asking, it closes the session in the activity *sessionClose* and migrates back to the relay node via the *goback* activity. After the relay node was reached it goes back to its root node. Finally, the survey result is deliverd in the *deliver* activity by inserting a `SURVEY` tuple in the local tuple space.

```
1: function survey (options) {
2:   this.root=null;
3:   this.relay=null;
4:   this.visited=[];
5:   this.chat=null;
6:   this.answer=null;
7:   this.token=null;
8:   this.timeout=10000;
9:   this.goback=false;
10:
11:  this.act = {
12:   init: () => {
13:     this.root=myNode()
```

```
14:   },
15:   percept: () => {
16:    if (test(['CHAT',_])) {
17:     // Found chat agent and chat app
18:     this.chat=myNode()
19:    }
20:   },
21:   search: () => {
22:    var connected=link(DIR.IP('%'))
23:    if (connected.length==1 && !this.relay) {
24:     // from root to relay
25:     this.relay=connected[0];
26:     moveto(DIR.NODE(this.relay))
27:    } else if (myNode() == this.relay) {
28:     // from relay to user device
29:     connected=filter(connected, (node) => {
30:      return (node != this.root) &&
31:             !contains(this.visited,node)
32:     })
33:     var next=random(connected);
34:     if (next) {
35:      moveto(DIR.NODE(next))
36:     } else {
37:      this.goback=true;
38:     }
39:    } else if (connected.length) {
40:     // from user back to relay
41:     this.visited.push(myNode())
42:     // Going back to relay
43:     moveto(DIR.NODE(connected[0]))
44:    } else sleep(500);
45:   },
46:   sessionOpen :  () => {
47:    // get a session token
48:    mark(['CHAT-SESSION',me(),60000],this.timeout);
49:    this.token=null;
50:    inp.try(this.timeout,['CHAT-TOKEN',me(),_], (reply) => {
51:     if (reply) this.token=reply[2];
52:    });
53:   },
54:   sessionClose : () => {
55:    mark(['CHAT-SESSION',me(),0],this.timeout);
56:   },
57:   ask : () => {
58:    // micro survey
59:    out(['CHAT-MESSAGE',me(),this.token,
60:         'Hello from the doctor from star '+this.root])
61:    out(['CHAT-QUESTION',me(),this.token,
62:         'What kind of person are you?',{
63:         choices:['quiet',
```

```
64:                    'restless',
65:                    'prudent',
66:                    'risky'],mutual:true}])
67:    // wait for answer or timeout
68:    inp.try(30000,['CHAT-ANSWER',me(),
69:            this.token,'What kind of person are you?',_], (t) => {
70:            this.answer=t;
71:    })
72:    },
73:    goback: () => {
74:     var next;
75:     if (myNode() == this.relay) {
76:      if (link(DIR.NODE(this.root))) next=this.root
77:     } else if (myNode() == this.chat) {
78:      if (link(DIR.NODE(this.relay))) next=this.relay
79:     }
80:     if (next) moveto(DIR.NODE(next));
81:      else kill()
82:    },
83:    deliver : () => {
84:     // Deliver the survey result
85:     out(['SURVEY','What kind of person are you?',
86:          this.answer,this.chat])
87:    },
88:    end : () => {
89:     kill()
90:    }
91:   }
92:   this.trans = {
93:    init : search,
94:    search : () => { return this.goback?goback:percept},
95:    percept : () => { return this.chat?sessionOpen:search },
96:    sessionOpen : () => { return this.token?ask:goback },
97:    ask : sessionClose,
98:    sessionClose : goback,
99:    goback : () => { return myNode()!=this.root?goback:deliver },
100:   deliver : end
101:  }
102:  this.next = init
103: }
```

*Ex. 20. Example of full featured computational explorer and survey agent (survey.js) that performs MCS on mobile platforms meeting a chat agent with user interface. It is assumed that the node origin as well as the JAM App are connected to an external relay node.*

## 10.6  Example of Chat Moderator Agent

The chat moderator agent can only be used on the JAM App (Cordova or Web browser version jamapp.html). A survey agent that want to talk with the user via the chat moderator must initiate a session (optionally with a access key) to get a session token. A session token is only granted for a specific time interval. The basic code of the chat moderator agent accessing an extended AIOS API (`message`, `question`, `notify`, `beep`) via the App is shown in Ex. 21 and the JAM App chat dialogue in Fig. 18. Client and moderator agent communicate entirely over the tuple space without knowing each other. Chat and user agent communicate entirely over the tuple space via four tuple patterns:

```
// From requesting agent to chat moderator:
['CHAT-MESSAGE',from,token,message]
['CHAT-NOTIFY',from,token,message]
['CHAT-QUESTION',from,token,question,action]
['CHAT-SESSION',from,time]
// From moderator to requesting agent
['CHAT-ANSWER',from,token,question,answer]
```

The chat agent waits for new session request in the activity *process*. If a session is opened, it waits for questions and messages from the current client agent. All communication is performed anonymously via the tuple space. Answers of the users are replied via the tuple space, too.

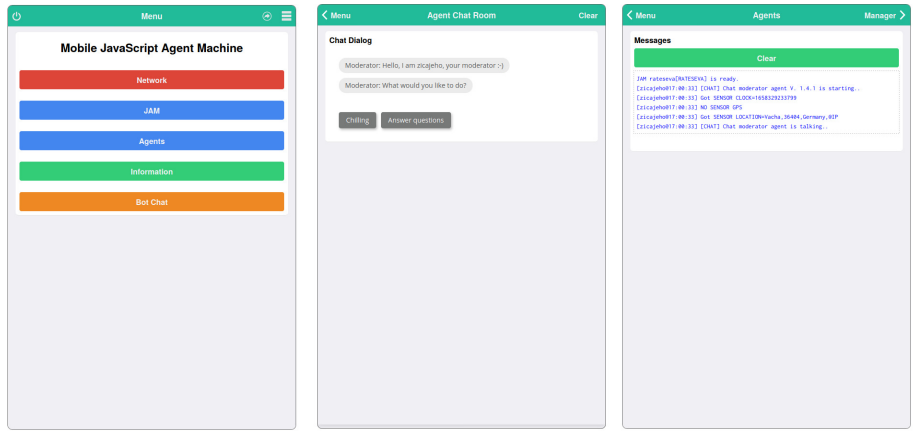

*Fig. 18. The JAM App chat dialogue controlled by the chat moderator agent (page view)*

```
 1:  function chat(options) {
 2:   // remember delay time
 3:   this.remember    = 2000;
 4:   // on or off
 5:   this.state       = 'off';
 6:   this.busy        = false;
 7:   // Current survey session? -> {from:string,stamp:number}
 8:   this.session     = null;
 9:   this.token       = 1001;
10:   this.sessionTime = 60000;
11:   // last chat dialog action
12:   this.last        = 0;
13:   // Default question-response timeout
14:   this.timeout     = 40000;
15:   // pending dialog requests
16:   this.dialogs     = [];
17:   this.idn         = 0;
18:
19:   this.act = {
20:    // Initialize the moderator agent
21:    init :  () => {
```

```
22:    var t;
23:    this.busy   = false;
24:    negotiate('CPU',10000000);
25:    this.idn=random(1000000);
26:    out(['CHAT',this.idn]);
27:  },
28:
29:  // Get feedback from user: chilling to chatting?
30:  talk: () => {
31:   message('Moderator','Hello, I am '+me()+', your moderator :-)');
32:   // Blocking operation (with timeout)
33:   this.state=null;
34:   this.busy=true;
35:   question('Moderator','What would you like to do?',[
36:       {text:'Chilling',value:'off'},
37:       {text:'Answer questions', value:'on'}
38:    ], (res) => {
39:     this.state=res;
40:     if (this.verbose) log('[CHAT] Got answer '+res+'.')
41:     if (res=='on') {
42:      notify('Now I will mediate questions from agents!');
43:      message('Moderator','Now I will mediate questions from agents!');
44:      message('Moderator','Hope you will answer their questions!');
45:      }
46:     this.busy=false;
47:    });
48:  },
49:
50:  process: () => {
51:   alt.try(500,[
52:      ['CHAT-MESSAGE',_,_,_],
53:      ['CHAT-NOTIFY',_,_,_],
54:      ['CHAT-QUESTION',_,_,_,_],
55:      ['CHAT-SESSION',_,_],
56:   ], (t) => {
57:    if (this.session && this.session.stop<time()) this.session=null;
58:    if (t) switch (t[0]) {
59:     case 'CHAT-MESSAGE':
60:      if (!this.session ||
61:          this.session.from!=t[1] ||
62:          this.session.token!=t[2]) return log(t);
63:
64:      this.dialogs.push({
65:       message: t[3],
66:       from:    t[1]
67:      });
68:      break;
69:
70:     case 'CHAT-NOTIFY':
71:      if (!this.session ||
```

```
72:            this.session.from!=t[1] ||
73:            this.session.token!=t[2]) return log(t);
74:
75:        notify(t[3]);
76:        beep();
77:        break;
78:
79:      case 'CHAT-QUESTION':
80:        if (!this.session ||
81:            this.session.from!=t[1] ||
82:            this.session.token!=t[2]) return log(t);
83:
84:        this.dialogs.push({
85:         question: t[3],
86:         from:     t[1],
87:         options:  t[4]
88:        });
89:        break;
90:
91:      case 'CHAT-SESSION':
92:       if (!this.session && t[2]>0) {
93:        this.session = {
94:         token :  this.token++,
95:         from  : t[1],
96:         stamp : time(),
97:         start : time(),
98:         stop  : time()+Math.min(this.sessionTime,t[2])
99:        };
100:        mark(['CHAT-TOKEN',this.session.from,this.session.token],1000);
101:       } else if (this.session && t[2]==0 && this.session.from==t[1]) {
102:        this.session=null;
103:       }
104:      break;
105:     }
106:   });
107: },
108:
109: refresh : () => {
110:   // refresh our chat tuple
111:   if (!ts(['CHAT',this.idn],600000)) {
112:    out(['CHAT',this.idn]);
113:   }
114: },
115: // process a doalog snippet
116: dodialog: () => {
117:   var action;
118:   if (this.busy) return;
119:   var d=this.dialogs.shift();
120:   if (d.message) {
121:    this.busy=true;
```

```
122:    message(d.from,d.message);
123:    this.busy=false;
124:   } else if (d.question) {
125:    this.busy=true;
126:    if (d.options.choices && d.options.mutual)
127:     action = d.options.choices;
128:    else if (d.options.choices && !d.options.mutual) {
129:     action = {
130:      placeholder : 'Select',
131:      multipleselect : true,
132:      value : '',
133:      button : {
134:       icon: 'check',
135:       label: 'OK'
136:      },
137:      options:d.options.choices
138:     };
139:    } else {
140:     action = {
141:      size:         14,
142:      value:        d.options.value||'',
143:      sub_type:     d.options.type || 'text',
144:      placeholder:  d.options.default||''
145:     };
146:    }
147:    question(d.from,d.question,action, (res) => {
148:     mark(['CHAT-ANSWER',d.from,this.session.token,d.question,res],10000);
149:     this.busy=false;
150:    },min(this.timeout,d.options.timeout||this.timeout));
151:   }
152:  },
153:
154:
155:  remember: () => {
156:   this.busy=false;
157:   sleep(this.remember)
158:  },
159:
160:  end: () => {
161:   kill()
162:  }
163: }
164:
165: this.trans = {
166:  init:talk,
167:  talk: () => {
168:   return this.state=='on'?refresh:remember
169:  },
170:  process: () => {
171:   if (this.dialogs.length) return dodialog;
```

```
172:    else return refresh
173:   },
174:   refresh : process,
175:   remember: talk,
176:   dodialog: refresh,
177:  }
178:
179:  this.next=init
180: }
```

*Ex. 21. Example of chat moderator agent (chat.js) that services MCS on platforms using the mobile JAM App*